\documentclass[lettersize,journal]{IEEEtran}
\usepackage{amsmath,amsfonts}
\usepackage{array}
\usepackage[caption=false,font=normalsize,labelfont=sf,textfont=sf]{subfig}
\usepackage{textcomp}
\usepackage{stfloats}
\usepackage{url}
\usepackage{verbatim}
\usepackage{graphicx}
\usepackage{xspace}
\usepackage{algorithm}
\usepackage{algpseudocode}
\usepackage{cite}
\usepackage{hyperref}
\usepackage{multirow}
\usepackage{booktabs}
\usepackage{dsfont}
\usepackage{amssymb}
\usepackage{bbm}
\usepackage{soul}
\usepackage[table,xcdraw]{xcolor} 
\def\BibTeX{{\rm B\kern-.05em{\sc i\kern-.025em b}\kern-.08em
    T\kern-.1667em\lower.7ex\hbox{E}\kern-.125emX}}
\usepackage{balance}
\begin{document}
\title{HERMES: \textbf{H}uman-to-Robot \textbf{E}mbodied Learning from Multi-Sou\textbf{R}ce \textbf{M}otion Data for Mobil\textbf{E} Dexterou\textbf{S} Manipulation}
\author{Zhecheng Yuan$^{1,2*}$, Tianming Wei$^{1,2*}$, Langzhe Gu$^{1,2}$, Pu Hua$^{1,2}$,\\ Tianhai Liang$^{1,2}$,  Yuanpei Chen$^{3}$, Huazhe Xu$^{1,2}$
\thanks{$^{*}$ indicates equal contribution. $^{1}$ Tsinghua University $^{2}$ Shanghai Qi Zhi Institute $^{3}$ Peking University 

Corresponding to huazhe\_xu@mail.tsinghua.edu.cn.
}

}

\markboth{}%
{How to Use the IEEEtran \LaTeX \ Templates}

\maketitle
\definecolor{darkgreen}{rgb}{0.09, 0.45, 0.27}
\definecolor{snsorange}{RGB}{255, 141, 98}
\definecolor{lyyblue}{RGB}{216,237,243}
\definecolor{lyyred}{RGB}{249,214,214}
\definecolor{lyydeepred}{RGB}{221,73,57}
\newtheorem{theo}{Theorem}
\newtheorem{definition}{Definition}
\newtheorem{assumption}{Assumption}
\newtheorem{lemma}{Lemma}
\definecolor{tablecolor}{RGB}{255,224,179}
\definecolor{ourorange}{RGB}{255,153,0}
\definecolor{tablecolor2}{RGB}{240, 220, 220}
\newcommand{\dd}[2]{$#1\scriptstyle{\pm#2}$}
\newcommand{\singledd}[1]{$#1$}
\newcommand{\singleddbf}[1]
{\cellcolor{lyyblue}$\mathbf{#1}$}

\newcommand{\ddbf}[2]
{\cellcolor{lyyred}$\mathbf{#1\scriptstyle{\pm#2}}$}
\newcommand{\blueddbf}[2]{%
  \cellcolor{lyyblue}%
  {\textcolor{lyydeepred}{$\mathbf{#1{\scriptstyle\,\pm\,}#2}$}}%
}

\newcommand{\ddbfblue}[2]{\cellcolor{tablecolor2}$\mathbf{#1\scriptstyle{\pm#2}}$}
\newcommand{\cc}[1]{$#1$}
\newcommand{\ccbf}[1]{\cellcolor{lyyred}$\mathbf{#1}$}
\newcommand{\best}[1]{\textcolor{lyydeepred}{\singleddbf{#1}}}
\newcommand{\ourshort}{HERMES\xspace}

\begin{figure*}[!t]
  \centering
  \includegraphics[width=1.0\linewidth]{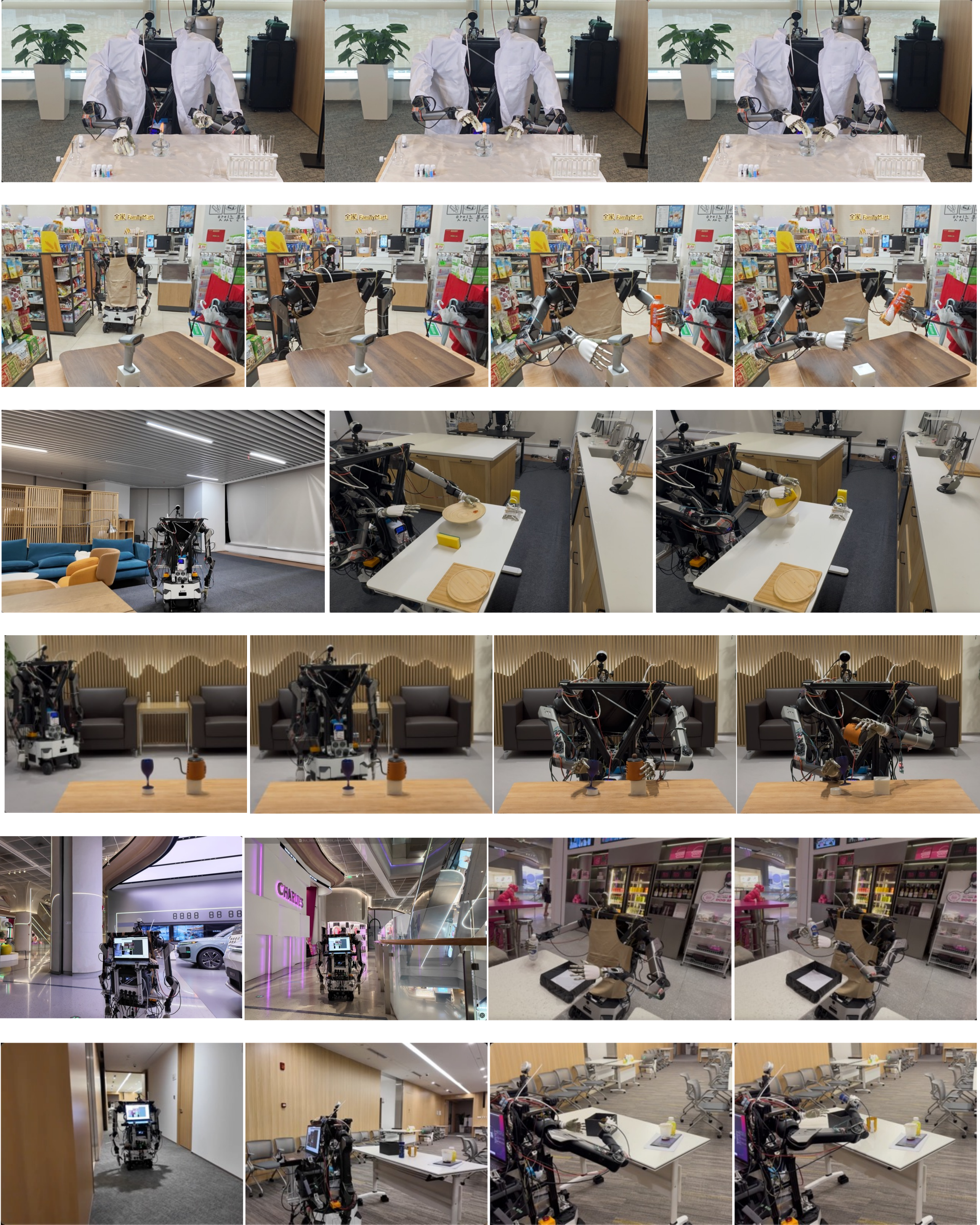}

  \caption{ \textbf{\ourshort exhibits a rich spectrum of mobile bimanual dexterous manipulation skills.} The robot is able to navigate over extended distances in both indoor and outdoor environments, and effectively execute a variety of complex manipulation tasks in unstructured, real-world scenarios, drawing upon behaviors learned from only one-shot human motion.}
  \label{fig:traj_vis}
\end{figure*}

\begin{abstract}
Leveraging human motion data to impart robots with versatile manipulation skills has emerged as a promising paradigm in robotic manipulation. Nevertheless, translating multi-source human hand motions into feasible robot behaviors remains challenging, particularly for robots equipped with multi-fingered dexterous hands characterized by complex, high-dimensional action spaces. Moreover, existing approaches often struggle to produce policies capable of adapting to diverse environmental conditions. In this paper, we introduce \ourshort, a human-to-robot learning framework for mobile bimanual dexterous manipulation. First, \ourshort formulates a unified reinforcement learning approach capable of seamlessly transforming heterogeneous human hand motions from multiple sources into physically plausible robotic behaviors. Subsequently, to mitigate the sim2real gap, we devise an end-to-end, depth image-based sim2real transfer method for improved generalization to real-world scenarios. Furthermore, to enable autonomous operation in varied and unstructured environments, we augment the navigation foundation model with a closed-loop Perspective-n-Point (PnP) localization mechanism, ensuring precise alignment of visual goals and effectively bridging autonomous navigation and dexterous manipulation. Extensive experimental results demonstrate that \ourshort consistently exhibits generalizable behaviors across diverse, in-the-wild scenarios, successfully performing numerous complex mobile bimanual dexterous manipulation tasks. Project Page:\url{https://gemcollector.github.io/HERMES/}. 
\end{abstract}


\section{Introduction}
\IEEEPARstart{A}{chieving} human-level dexterity for robots has long been a central challenge in robotic research. The prospect of bimanual robotic systems with dexterous hands that mirror human sensorimotor dexterity holds the promise of seamlessly integrating robots into daily human activities and environments. Despite notable progress, how to capitalize on the abundance of available human data and develop algorithms suited to intricate and high-precision dexterous manipulation remains underexplored.

Humans continuously generate diverse bimanual manipulation data, inherently serving as natural guidance for robots to emulate human-like behaviors. Several previous studies~\cite{zhou2025you,kim2025uniskill,lum2025crossing,dan2025x,wangmimicplay} have attempted to extract trajectories of human hands and manipulated objects from video data, subsequently applying them to robotic manipulation tasks. Nevertheless, these methods have predominantly targeted robots equipped with simple parallel gripper end effectors, failing to generalize effectively to dexterous hands due to the vastly greater complexity of action space. Despite recent advances that utilize kinematic retargeting approaches to produce human-like robotic motions~\cite{qiu2025humanoid,qin2023anyteleop,yang2025egovla,shaw2024learning,shaw2023videodex}, these approaches still fall short in achieving physically-aware pose retargeting and bridging the embodiment gap to derive feasible robot actions capable of successfully accomplishing the intended tasks. A critical limitation lies in omitting the modeling of interactions between robotic hands and manipulated objects, a fundamental component of manipulation tasks. Consequently, neglecting these interactions undermines the robot's ability to fully understand and adapt to the dynamics of manipulation scenarios.

Therefore, in an attempt to address the aforementioned challenges, recent approaches have begun leveraging Reinforcement Learning~(RL) paradigms~\cite{chenobject,li2025maniptrans,mandi2025dexmachina}, allowing robots to autonomously explore feasible motion strategies under the guidance of kinematic reference trajectories. These methods commonly design general reward functions encompassing object tracking, hand configurations, and collision dynamics. Maximizing such rewards drives the robot toward successful execution of complex manipulation tasks. Nonetheless, existing works~\cite{mandi2025dexmachina,li2025maniptrans,chenobject} typically draw on limited human motion data sources and some have not transferred the trained robot behaviors to the physical world. Such limitations not only hinder the evaluation of whether the learned policies exhibit behaviorally plausible performance in the real world, but also preclude the integration of sim2real methodologies necessary for robust policy control and for deployment across various environmental conditions. Furthermore, current methods for sim2real transfer of bimanual dexterous manipulation predominantly rely on explicit extraction of object and robot state information~\cite{lin2024learning,chenobject,chen2023sequential,lin2024twisting}, thus failing to achieve end-to-end visual learning. This limitation inherently confines learned policies to specific fixed setups, significantly hindering their adaptability to diverse scenarios.

Motivated by these challenges, we propose \ourshort, a versatile human-to-robot embodied learning framework tailored for mobile bimanual dexterous hand manipulation. \ourshort offers the following three advantages:
\begin{itemize}
    \item \textbf{Diverse sources of human motion}: Our framework supports several human motion sources, including teleoperated simulation data, motion capture~(mocap) data, and raw human videos. We also provide corresponding approaches for data acquisition, enabling \ourshort to efficiently transform varied human motion data into robot-feasible behaviors through RL. Furthermore, these tasks share a uniform set of reward terms, obviating the necessity of designing intricate and task-specific reward functions. In contrast to the methods that depend on collecting a large amount of demonstrations,  we can achieve generalizable policy by augmenting a single reference human motion trajectory coupling with RL training.
    \item \textbf{End-to-end vision-based sim2real transfer}: \ourshort facilitates robust vision-based sim2real transfer by employing DAgger distillation, which converts state-based expert policies into vision-based student policies. Moreover, we introduce a generalized, object-centric depth image augmentation and hybrid control approach, effectively bridging the perception and dynamic sim2real gap.
    \item  \textbf{Mobile manipulation capability}: Our method endows robots with mobile manipulation skills. Building upon ViNT~\cite{shah2023vint}, we develop a RGB-D based module for precise localization wherein the task is modeled as a Perspective-n-Point~(PnP) problem and addressed through an iterative process. This ensures seamless integration with subsequent manipulation tasks and unlock the policy’s capacity to operate autonomously across a broad spectrum of real-world environments.
    
\end{itemize}

With the integration of these capabilities, \ourshort is empowered to execute a wide range of complex or long-horizon mobile bimanual dexterous manipulation tasks across varied and unstructured real-world environments. We conduct extensive experiments in both simulation and real-world settings. The experimental results demonstrate that \ourshort achieves superior performance with high task success rates and high sample efficiency. The learned policies can not only successfully transfer to the physical world but also exhibit generalization capabilities. Furthermore, the navigation component of \ourshort demonstrates precise localization ability, facilitating effective deployment of trained manipulation policies in diverse in-the-wild scenarios.

\section{Related Work}
\subsection{Dexterous Manipulation from Human Demonstrations} 
By harnessing human motion as the fuel of robot data, robots can acquire natural and versatile behaviors~\cite{argall2009survey,mandikal2022dexvip,shaw2023videodex,ye2023learning,qin2022dexmv,luo2024omnigrasp,ze2025twist,he2024learning}. In contrast to the typically scarce and costly teleoperation datasets, human motion data offers a more abundant and economically accessible resource for training robots. Recent advancements~\cite{zhou2025you,kim2025uniskill,lum2025crossing,dan2025x,wangmimicplay} have seen numerous studies leveraging human videos to extract features that assist in downstream policy learning. However, these efforts predominantly focus on transferring knowledge to robots with parallel-jaw end-effectors, often overlooking the necessity of bridging the embodiment gap and considering the complexity of high action space. In contrast, dexterous hands present greater challenges in modeling grasping postures and interactions with objects. Some recent works~\cite{kareer2024egomimic,hoque2025egodex,qiu2025humanoid} have employed customized, high-precision equipment to collect egocentric human videos for training robot policies, yet they neglect to model the complexity of hand-object interactions. Additionally, several recent studies~\cite{liu2025dextrack,chenobject,li2025maniptrans} utilize reinforcement learning to translate human hand motions into robotic behaviors, but these approaches are typically limited to simple, single-object tasks or lack closed-loop sim2real implementations. With a suite of innovative designs, \ourshort overcomes these shortcomings and equips the robot to perform various challenging, high–degree-of-freedom bimanual dexterous manipulation tasks. Moreover, we effectively transform diverse types of human motion into deployable robot policies.

\subsection{Vision-based Sim2real Manipulation}
Sim2real has achieved notable advancements in locomotion, with policies trained in simulation successfully deployed on quadruped or humanoid robots to perform agile motions~\cite{zhuang2023robot,zhuanghumanoid,rudin2025parkour,allshire2025visual,lei2024unio,qi2023hand}. In terms of manipulation, recent research has increasingly focused on leveraging simulation to generate or augment training data to obtain deployable visuomotor policies~\cite{wang2024cyberdemo,maddukuri2025sim,yuanlearning,huagensim2,wanggensim,xu2025dexsingrasplearningunifiedpolicy,zhang2025robustdexgrasp}. 
However, manipulation tasks, particularly those involving bimanual dexterous manipulation, present greater challenges for sim2real transfer due to the presence of high-frequency, fine-grained visual information.
Several approaches~\cite{lin2024learning,lin2024twisting,chen2023sequential,lin2025sim} have utilized depth cameras to extract object poses, shapes, and other relevant information, combining these with proprioceptive state data to facilitate sim2real transfer. While these methods can accurately obtain robot and object state information, the resulting policies often lack the ability to visually perceive environmental and object changes, which in turn forces manual specification of numerous state variables. Another line of works~\cite{yuanlearning,bjorck2025gr00t,singh2024dextrah} employ extensive domain randomization to train generalizable visual policies. However, achieving diversity in such randomization schemes typically involves labor-intensive engineering efforts. Recent studies~\cite{qureshi2024splatsim,li2024robogsim} have explored the use of Gaussian splatting~\cite{kerbl20233d} to achieve photorealistic rendering, but this approach requires manual scanning of entire scenes and training tailored to specific robot embodiments and environments, which limits scalability. 

Depth images inherently preserve object shape and spatial structure, enabling efficient sim2real transfer without the need to overcome variations in textures~\cite{dalal2024local,lumdextrah}. For instance, DextrAH-G~\cite{lumdextrah} utilizes depth maps for sim2real transfer but necessitates manually designing finely tuned and specific noise. Here, we also leverage depth images to facilitate egocentric sim2real transfer in an end-to-end fashion. In contrast to DextrAH-G, our approach adopts a more generalizable depth image augmentation strategy, which eliminates the need for camera-specific noise modeling. By applying our depth image processing method, we not only achieve strong semantic alignment between simulated and real-world observations but also obtain robust and generalizable visuomotor policies capable of handling complex manipulation tasks.

\subsection{Mobile Manipulation}
Mobile manipulation tasks further compound the challenges of robotic manipulation~\cite{yenamandra2023homerobot,fu2024mobile,huang2023skill,Liu_2024,zhi2025closedloopopenvocabularymobilemanipulation,wu2023tidybot}. The robot should typically complete a navigation phase before proceeding with the manipulation task. Many recent studies~\cite{Liu_2024,fang2023anygrasp,langsam2023,zhi2025closedloopopenvocabularymobilemanipulation} have adopted a modular mobile manipulation framework and utilized pre-built maps for robot mobile manipulation. OK-Robot~\cite{Liu_2024} uses a 3D map of the room built with iPhone and leverages an open-vocabulary object detector to perform open-vocabulary object navigation with the map. It also adopts AnyGrasp~\cite{fang2023anygrasp} combined with LangSam~\cite{langsam2023} to perform open-vocabulary grasping in the real world. COME-robot ~\cite{zhi2025closedloopopenvocabularymobilemanipulation} utilizes a global object map built by the robot for global-level perception. It leverages GPT-4V~\cite{gpt4v} to perform target object perception and task planning. However, these map-based methods are fundamentally constrained to small indoor spaces, as they become computationally intractable for large-scale environments like outdoor areas. They also struggle in feature-deprived settings or mixed indoor-outdoor navigation scenarios, where reliable localization becomes particularly challenging.

Another line of research explores end-to-end training paradigms that unify mobile navigation and manipulation into a single framework~\cite{jiang2025behavior,xiong2024adaptive,meng2025aim,yang2024harmonic}. However, such approaches not only demand a larger number of demonstrations or task-specific training, but also fail to leverage prior knowledge from navigation systems, thus constraining them to short-horizon tasks. To equip visuomotor policies with in-the-wild navigation capabilities, we incorporate the image-goal navigation foundation model that relies exclusively on RGB inputs and enables generalizable navigation, while being seamlessly integrated with the trained manipulation policy.

\section{System Design}
We build a mobile bimanual robot equipped with two dexterous hands. To enable large-scale training in simulation, we also construct corresponding high-fidelity simulation models that accurately replicate the physical characteristics and kinematic structures of the real-world system.

\begin{figure}[t]
  \centering
  \vspace{-20pt}
  \includegraphics[width=1.0\linewidth]{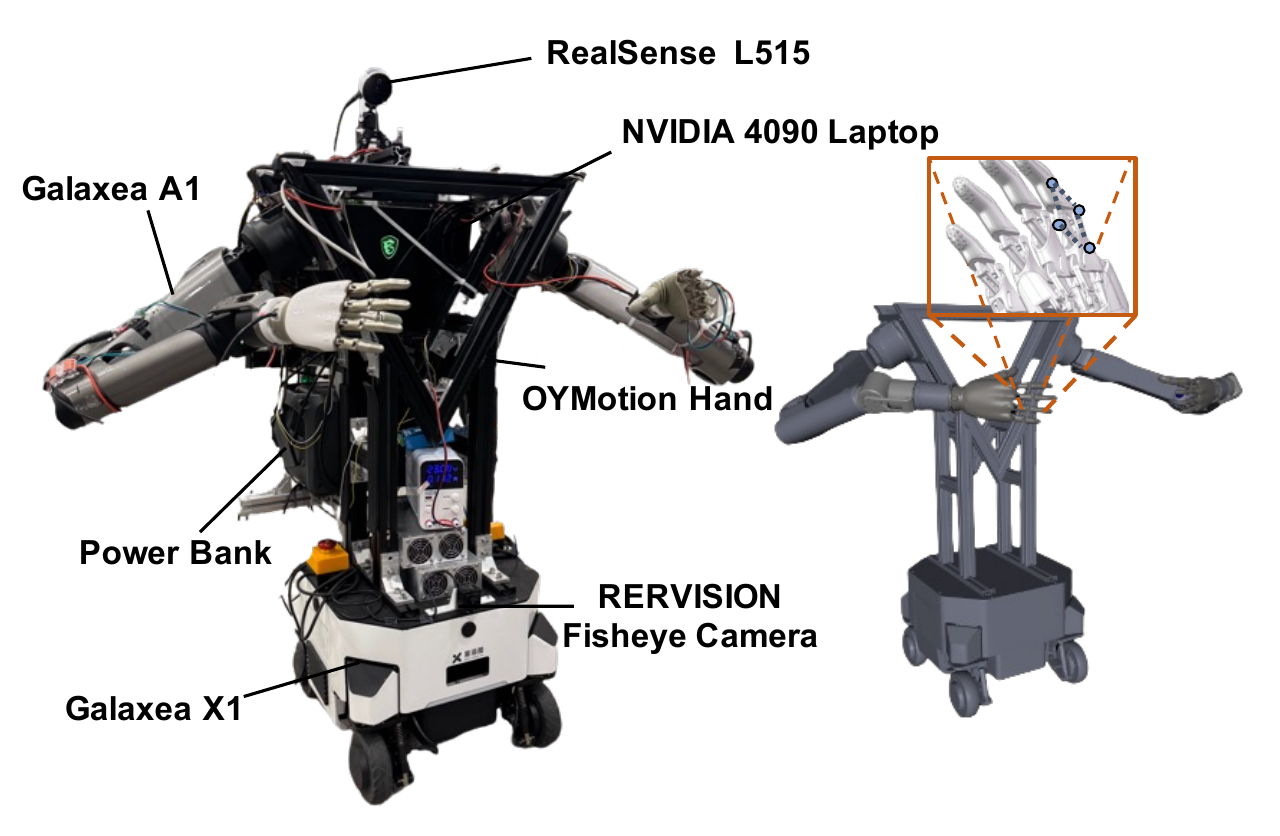}
  \vspace{-15pt}
  \caption{\textbf{System Design.} We construct a unified setup of mobile bimanual robots equipped with dexterous hands in both simulation and the real world. Through high-fidelity simulation, this robotic platform is capable of enabling sim2real transfer across a wide range of complex manipulation tasks.}
  \label{fig:system}
\end{figure}

\subsection{Hardware Design}
As shown in Figure~\ref{fig:system}, our robotic system is constructed by integrating an X1 mobile base, two 6-DoF Galaxea A1 arms, and two OYMotion 6-DoF dexterous hands. The torso part is assembled using lightweight aluminum alloy tubing to ensure structural rigidity. Additionally, we use a laptop with an NVIDIA RTX 4090 GPU for supporting ROS control and network inference. Regarding the camera, we adopt the RealSense L515 to capture RGBD observations and the RERVISION Fisheye camera for navigation. 

\subsection{Simulation Design}
We employ both the MuJoCo~\cite{todorov2012mujoco} and MJX simulation platforms~\cite{MJX} to construct training environments tailored for our mobile bimanual robotic system. The kinematic structure of the robot is carefully modeled, and relevant dynamic parameters are configured to ensure stable behavior in the simulation. The actuation range of each joint is configured to match that of the physical robot. For the dexterous hands, whose fingers contain passive joints not directly actuated by motors, we leverage MuJoCo’s capabilities for modeling closed chain mechanisms to faithfully simulate these passive DoFs, which achieves a high-fidelity reproduction of the hand's physical interactions. In contrast to the conventional approaches of simulating inter-link motion relationships through mimic joints or tendon-based mechanisms,  we utilize the equality constraint feature in MuJoCo to directly construct linkage structures as defined in the original CAD models. This formulation affords a more precise representation of the motion dependencies among passive DoFs, leading to a more faithful simulation of constrained multi-link dynamics.

Furthermore, to enhance the stability of interactions between the robot and manipulated objects, we restructure the collision modeling scheme. Rather than performing collision detection based directly on the original mesh representations of the objects and the hand, we approximate their geometries using primitive shapes. This abstraction facilitates more granular and stable collision computation. The Inverse Kinematics~(IK) control of the robotic system is implemented using the Mink library~\cite{ZakkaMink2025}.

\begin{figure*}[t]
  \centering
  \includegraphics[width=1.0\linewidth]{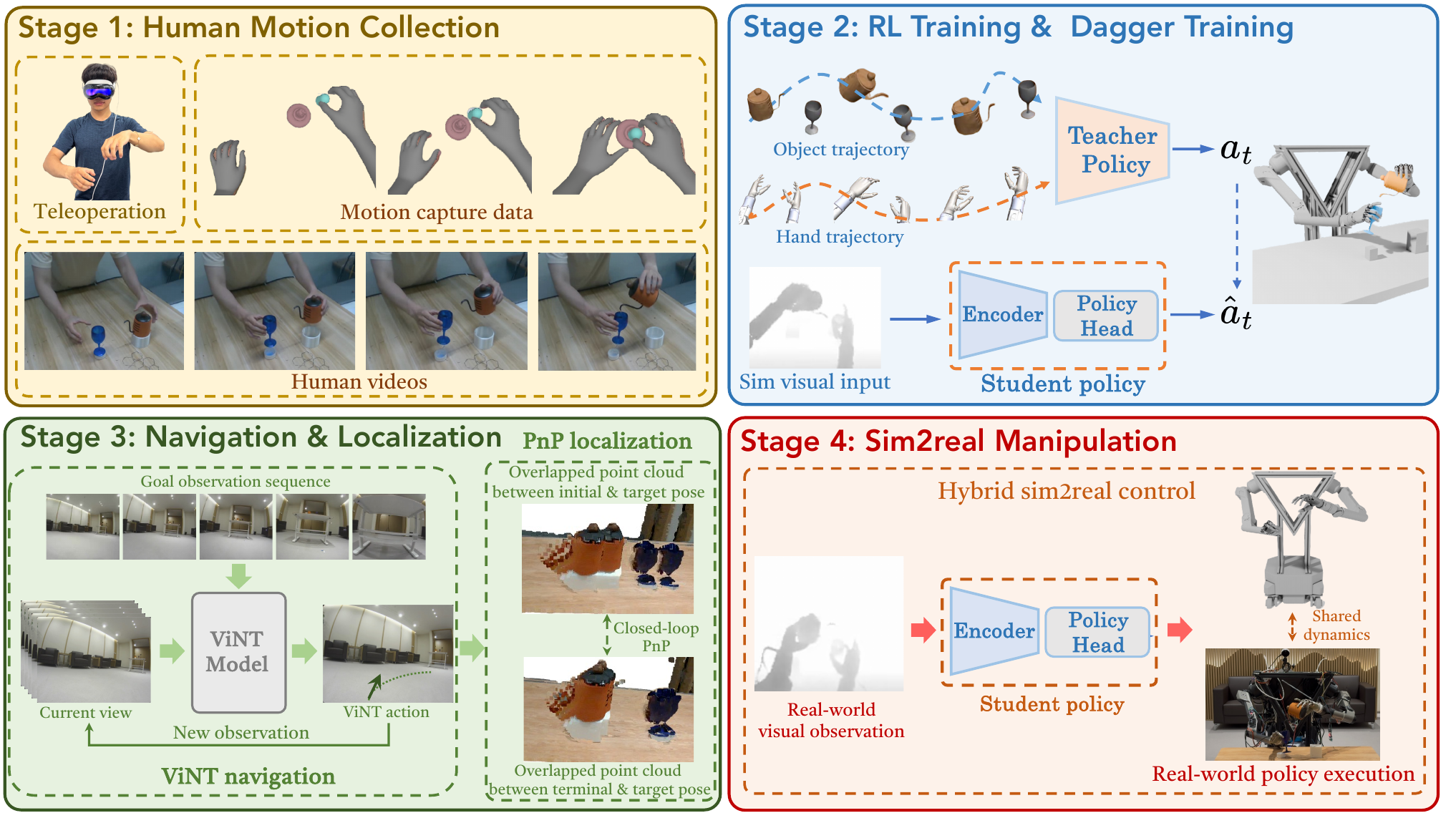}
  \vspace{-15pt}
  \caption{ \textbf{The main pipeline of \ourshort.} \ourshort comprises a four-stage pipeline for achieving mobile bimanual dexterous manipulation through sim2real transfer. First, we acquire a one‑shot human demonstration drawn from diverse sources. Then, in stage 2, we train a state-based RL teacher policy, then apply DAgger to distill it into a vision‑based student policy. Following this, \ourshort execute long‑horizon navigation using ViNT, followed by closed-loop PnP to finely adjust the robot’s pose and achieve precise alignment in stage 3. Once localization is achieved, the student policy is deployed in a zero‑shot fashion directly in the real world.}
\label{fig:method}
\end{figure*}
\section{Reinforcement Learning Method}
\subsection{Task Formulation}
We define each task under the framework of goal-conditioned reinforcement learning. Our goal is to enable the robot to acquire the capability to convert the desired kinematic motion trajectory into physically executable robot behaviors. The tasks are formulated as the Markov Decision
Process~(MDP) $\mathcal{M}=(\mathcal{S}, \mathcal{A}, \mathcal{T}, \mathcal{R}, \gamma, \mathcal{G}$) , where $\mathcal{S}$ is the state space, $\mathcal{A}$ is the action space, $\mathcal{T}$ is the transition function, $\mathcal{R}$ is the reward, $\gamma$ is the discount factor, $\mathcal{G}$ is the reference trajectory. For goal-conditioned reinforcement learning,  the state $s \in \mathcal{S}$  includes the proprioception information $s^{p}$ and the goal state $s^{g}$ from $\mathcal{G}$, and the reward function at timestep $t$ is defined as $r_{t}=\mathcal{R}(s^{p}_{t}, s^{g}_{t})$. Under such a formulation, the policy is able to acquire complex and fine-grained behaviors from diverse reference trajectories without relying on labor-intensive reward engineering.

\subsection{Collect One-shot Human Motion}
To validate the effectiveness and robustness of \ourshort, we employ three distinct sources of human motion: teleoperation in simulation, motion capture data obtained from public datasets, and hand-object poses extracted from raw videos. Moreover, by leveraging merely a single human reference trajectory in conjunction with RL training, we are able to derive the generalizable robot policy without the need for collecting extensive demonstrations.

\textbf{Teleoperation in simulation:} We provide access to the pre-configured simulation that enables direct teleoperation of the robot for collecting demonstrations. The Apple Vision Pro is utilized to extract hand poses and arm movements, with data captured at a frequency of 75 Hz.

\textbf{Mocap data:} In contrast to direct teleoperation in simulation, retargeting mocap data to robotic hands presents significant challenges due to the embodiment gap between human and robotic hand structures. This discrepancy renders the retargeted trajectories from mocap data unsuitable for direct replay in simulation. Consequently, RL is often employed to enable robots to learn the desired behaviors from reference trajectories. In our study, we acuqire human motion capture data from the OakInk2~\cite{zhan2024oakink2} dataset. 

\begin{figure}[h]
  \centering
  \includegraphics[width=1.0\linewidth]{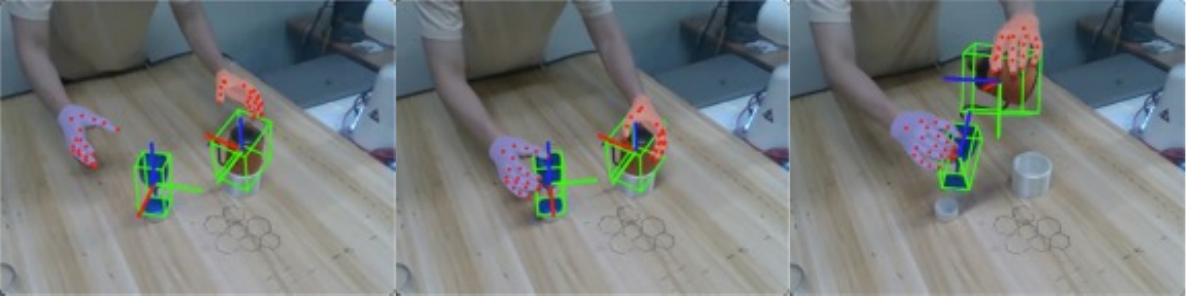}
  \vspace{-15pt}
  \caption{\textbf{Pose extraction from videos.} We utilize FoundationPose to extract the pose trajectories of multiple objects and employ WiLoR to capture the poses of both hands along with the positions of their finger joints.}
  \label{fig:fp}
\end{figure}
\begin{figure}[h]
  \centering
  \includegraphics[width=1.0\linewidth]{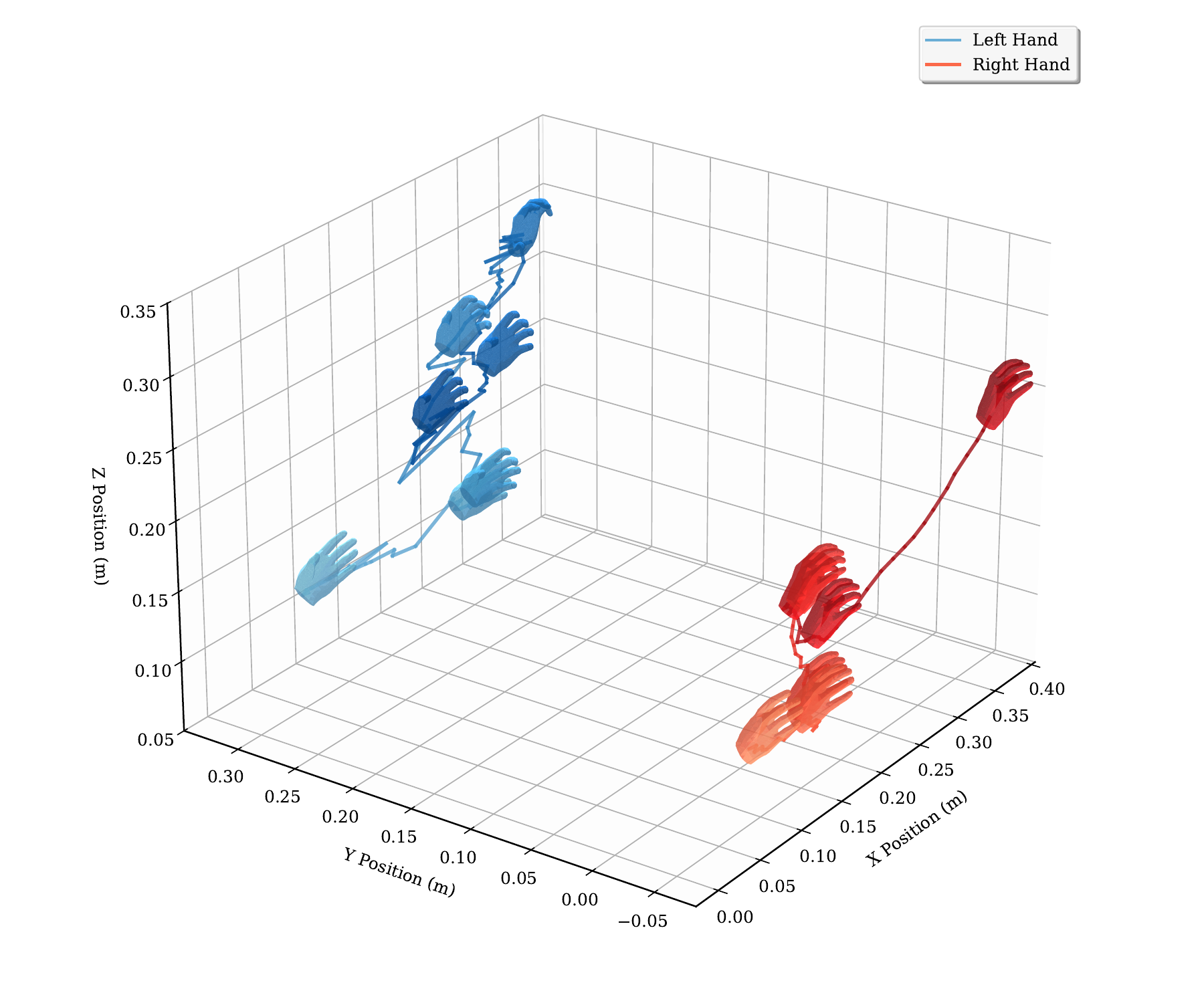}
  \vspace{-15pt}
  \caption{ \textbf{The visualization of hand motion trajectory.} We utilize WiLoR along with a PnP algorithm to precisely transform the estimated hand poses into the robot’s frame.}
  \label{fig:handtraj}
\end{figure}
\textbf{Extracted arm and hand poses from videos:} 
Leveraging video data holds considerable promise for unlocking vast quantities of information to facilitate robot learning. To this end, we also provide a pipeline for extracting human hand poses and object trajectories directly from raw video. To acquire the hand poses, we first employ WiLoR~\cite{potamias2024wilor} to detect the hands in each video frame and extract 2D hand keypoints along with their corresponding 3D counterparts. We then select a relatively stable subset of keypoints for the subsequent estimation, specifically those located at the wrist and the metacarpophalangeal joints. The spatial translation of the wrist in the camera coordinate system is estimated by solving a Perspective-n-Point~(PnP) problem~\cite{li2012robust} based on the 2D-3D correspondences, while the palm's orientation is derived by fitting a plane to the selected 3D keypoints. The extraction results are shown in Figure~\ref{fig:handtraj}. Regarding the manipulated objects, we employ FoundationPose~\cite{wen2024foundationpose} to estimate the object poses directly from video frames, and utilize ARCode~\cite{arcode2022} scanning to reconstruct the object mesh. By leveraging the aforementioned procedures, we can align the hand and object poses extracted from the video with the robot's frame to facilitate the subsequent learning process.   

\textbf{Synthesize multiple trajectories:} To obtain a more generalizable policy, we perform the trajectory augmentation for the one-shot human motion reference by randomizing positions and orientations of the objects in a predefined range. The hand and object poses across the augmented trajectories are transformed as follows:
\begin{equation}
\hat{\mathbf{A}}^{\mathrm{pose}}\left[\tau_k\right]=\mathbf{T}^{\mathrm{trans}} \cdot\mathbf{A}^{\mathrm{pose}}\left[\tau_k\right] .
\end{equation}
For any given frame $k$ in the trajectory $\tau$, we apply a transformation matrix $\mathbf{T}^{\mathrm{trans}}$ to alter its pose, where $\mathbf{A}^{\mathrm{pose}}$ may represent either the object pose or the hand pose. By editing the reference trajectory, we enable spatial generalization from a single human motion demonstration, obviating the need to manually collect large numbers of teleoped demonstrations.

Upon obtaining synthesized object and hand trajectories from various data sources, we initially employ DexPilot~\cite{handa2020dexpilot}, a popular retargeting method to map the captured human hand poses onto corresponding robot hand configurations. Subsequently, reinforcement learning is leveraged to refine and adapt the initialized robot behaviors.

\subsection{Generalizable Reward Design for Manipulation}
Standard reinforcement learning typically relies on hand-crafted reward functions tailored to each specific task. However, designing such complicated reward structures often impedes scalability and usability, particularly for the dexterous hand. To alleviate this issue, we leverage one-shot human demonstration combined with a generalizable reward formulation so that one unified reward function can be reused across tasks and simplify the specific design of challenging, long-horizon manipulation tasks. Recent advancements in locomotion~\cite{luo2023perpetual,peng2018deepmimic,he2025asap,allshire2025visual} have predominantly adopted such design paradigms for training humanoid or quadruped robots. However, manipulation tasks introduce additional complexities, as robots must not only perceive their own proprioception states but also effectively interact with external objects. Therefore, we propose a generalizable reward framework tailored for manipulation tasks, capable of being applied across diverse tasks. 

Specifically, we design the following three reward terms:

\begin{figure}[t]
  \centering
  \includegraphics[width=1.0\linewidth]{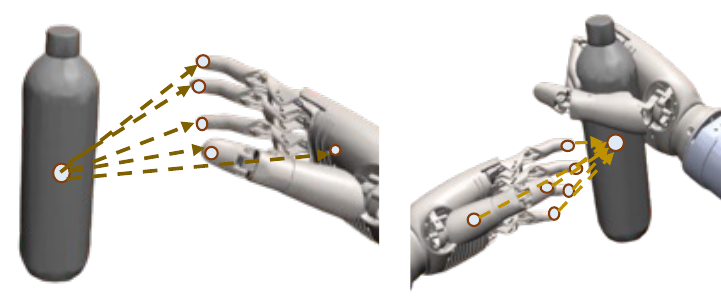}
  \vspace{-15pt}
  \caption{ \textbf{Object-centric distance chain.} This reward term is computed by tracking the temporal variations of vectors formed between the object center and each fingertip as well as the palm of both hands.}
  \label{fig:distance-chain}
\end{figure}
\textbf{Object-centric distance chain:}  
Capturing the dynamic spatial relationships between the human hands and the object stands as a pivotal factor in enabling the policy to acquire fine-grained hand-object interaction skills. As illustrated in Figure~\ref{fig:distance-chain}, we designate the coordinates of the fingertips and palm of the hand, along with  the center of the object's collision mesh, as keypoints. By modeling the temporal variation of vectors between these keypoints, we formulate the following reward function:

\begin{equation}
r_{\text{chain}} =
\begin{cases}
\exp\left\{- \frac{1}{n} \sum_{i=1}^n \left\| \vec{r}_{\text{ref}}^{(i)} - \vec{r}^{(i)} \right\| \right\}, & \text{if } N_{\text{contact}} \geq N_{\text{num}} \\
0, & \text{otherwise},
\end{cases}
\end{equation}

where ${\vec{r}}^{(i)}$ is the vector from object center to the fingertip or palm. Furthermore, we incorporate contact information into this reward term.  Specifically, during the computation of the distance chain, we also evaluate the number of contact points between the fingertips and palms of both hand mesh $\mathbf{C}_{\text{hand}}$ and the object's collision mesh $\mathbf{C}_{\text{obj}}$. This reward component is activated only when the number of contact points $N_{\text{contact}}$ exceeds a predefined threshold $N_{\text{num}}$, ensuring that the policy attends to physically meaningful hand-object interactions.
\begin{equation}
    N_{\text{contact}} = \frac{1}{2}\sum_{j}\sum_i \mathbbm{1}(\mathbf{C}_{\text{hand}}^{i}, \mathbf{C}_{\text{obj}}^{j}),
\end{equation}

where $\mathbbm{1}(\cdot, \cdot)$ is the indicator function that evaluates whether a collision occurs between the hand and the object. In addition, we also encourage the robot to favor joint positions in the retargeted trajectories during exploration.

\textbf{Object trajectory tracking:}
For manipulation tasks which adopt human motion, a critical indicator of policy success lies in its ability to track and follow the desired object trajectory. To this end, we introduce an additional reward component that explicitly aligns the policy’s behavior with the target object’s trajectory:

\begin{equation}
r_{\text{obj}} = \exp\left(-k_1 \cdot \left\| \mathbf{p}_{\text{obj}} - \mathbf{p}_{\text{ref}} \right\|^2 - k_2 \cdot \left( d_{\text{quat}}(\mathbf{q}_{\text{obj}}, \mathbf{q}_{\text{
ref
}}) \right)^2\right)
\end{equation}

where $k_{1}$, $k_{2}$ are the coefficients corresponding to the position and orientation terms in the $r_{\text{obj}}$.  $\mathbf{p}_{\text{obj}}$ and $\mathbf{q}_{\text{obj}}$ represent the current position and orientation of the object, while $\mathbf{p}_{\text{ref}}$ and $\mathbf{q}_{\text{ref}}$ denote the position and orientation along the reference trajectory. The term $d_{\text{quat}}$ measures the distance between two quaternions.

\textbf{Power penalty:} In order to enhance the smoothness of policy execution, we design a power penalty reward to alleviate the jittering actions:

\begin{equation}
r_{\text{penalty}} = -\lambda \sum_{j \in \mathcal{J}_{\text{hand}}} \left| f_j \cdot \dot{q}_j \right|,
\end{equation}

where $\lambda$ is the coefficient of $r_{\text{penalty}}$. $\mathcal{J}_{\text{hand}}$ denotes the set of all hand joints. $f_{j}$ is the actuation force applied at joint $j$, and $\dot{q}_j$ is the velocity of joint $j$.

By integrating all these reward components, the policy is endowed with the capacity to tackle a wide spectrum of challenging and diverse manipulation tasks.

\subsection{Residual Action Learning}
 While human trajectories provide coarse guidance on arm and hand movements throughout task execution, they often lack the precision required for accurate object interactions. To address this, we design distinct learning strategies for the arm and hand. For the arm, we decompose actions into coarse and fine components. At each timestep $t$,  the coarse action $\boldsymbol{a_{g}}$ is directly derived from the human trajectory, offering a general movement direction. The fine component $\Delta_{\boldsymbol{a_{f}}}$ is predicted by a learned network $\Delta_{\boldsymbol{a_{f}}}=\boldsymbol{\pi(s)}$, refining the motion to ensure precise control. Hence, the final form of arm action $\boldsymbol{a_{arm}}$ is $\boldsymbol{a_{arm}}=\Delta_{\boldsymbol{a_{f}}}+\boldsymbol{a_{g}}$. Regarding the hand part, due to inaccuracies in retargeting human demonstrations, we entirely adopt the network output $\boldsymbol{a_{hand}}$  to model the interaction behaviors with objects. This hybrid approach allows the policy to capture the overarching structure of human motion while learning the nuanced adjustments necessary for effective and precise manipulation for robots. 

In addition, given the high-dimensional action space and inherent complexity of bimanual dexterous hand tasks, we incorporate the early termination strategy to curtail inefficient exploration~\cite{peng2018deepmimic}. Meanwhile, to enable the acquisition of an initial viable behavior, we disable object collisions during the early stages of training, allowing the policy to first focus on learning the approximate motion trajectory.

\subsection{Reinforcement Learning Algorithm}
We implement two distinct reinforcement learning algorithms. DrM~\cite{xudrm}, an off-policy method, leverages a dormant ratio mechanism~\cite{sokar2023dormant} to enhance exploration capabilities and demonstrates high sample efficiency. We adapt the network architecture and take the state observation as inputs for state-based RL training. DrM training for all tasks is conducted in the MuJoCo simulation environment. Concurrently, we also instantiate the tasks in the MJX platform, which supports GPU-accelerated parallel training, and employ the PPO algorithm to significantly reduce the overall wall-clock training time. More details can be found in Appendix~\ref{sec:state-rl}.

\section{Sim-to-real Transfer}
The training of state-based RL policies typically relies on privileged information which is not accessible in real-world deployment scenarios. Consequently, it is imperative to distill the state-based policy into a visual policy for achieving sim2real transfer.

\begin{figure*}[t]
  \centering
  \includegraphics[width=1.0\linewidth]{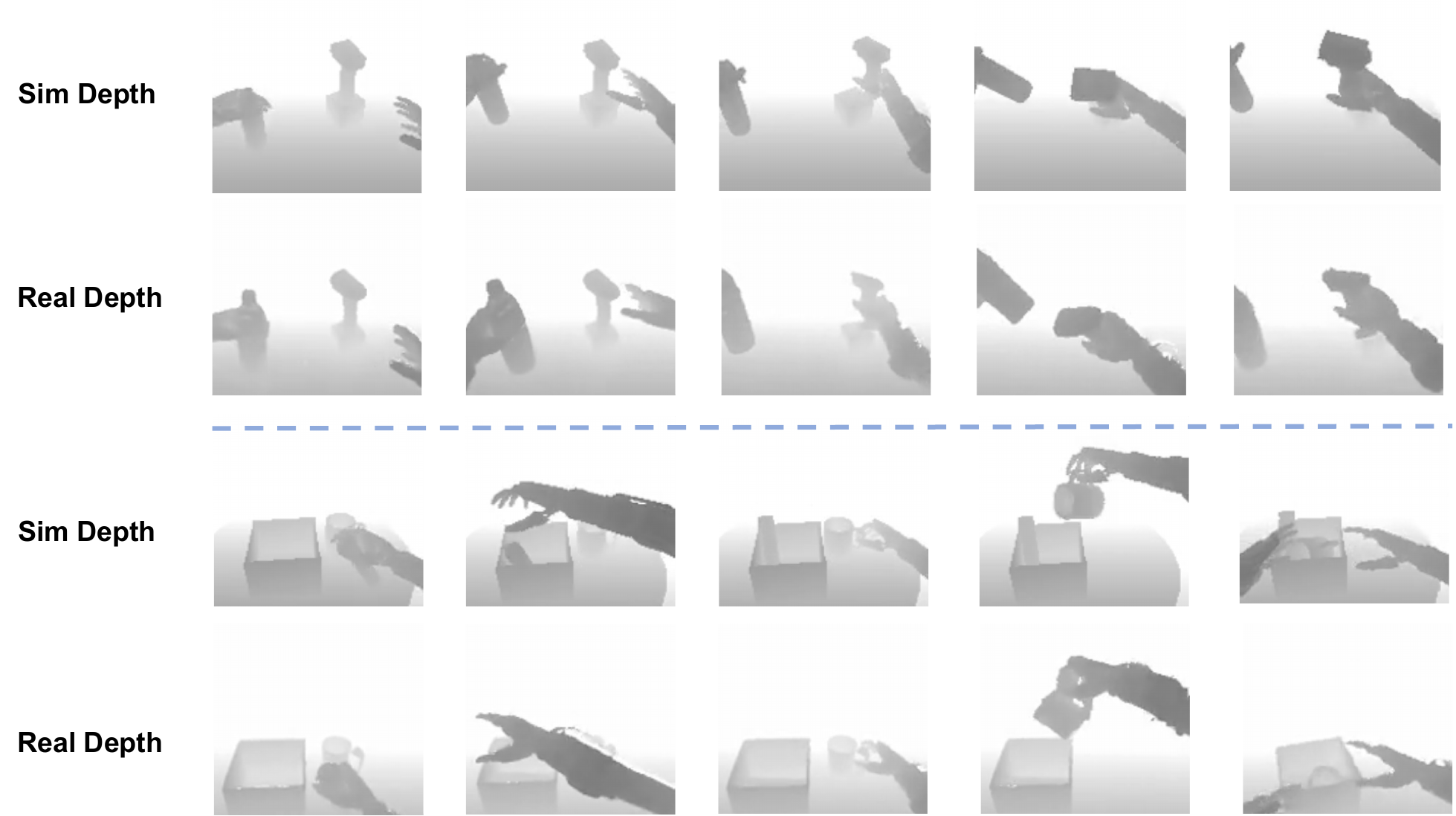}
  \vspace{-15pt}
  \caption{\textbf{Depth image visualization.} We present a visual comparison between simulated and real-world depth maps across two different tasks. Notably, after applying our preprocessing pipeline, the depth representations of the hand and object exhibit a strong semantic correspondence, highlighting the efficacy of \ourshort in bridging the sim2real gap.}
  \label{fig:compare_depth}
\end{figure*}

\begin{figure}[h]
  \centering
  \includegraphics[width=1.0\linewidth]{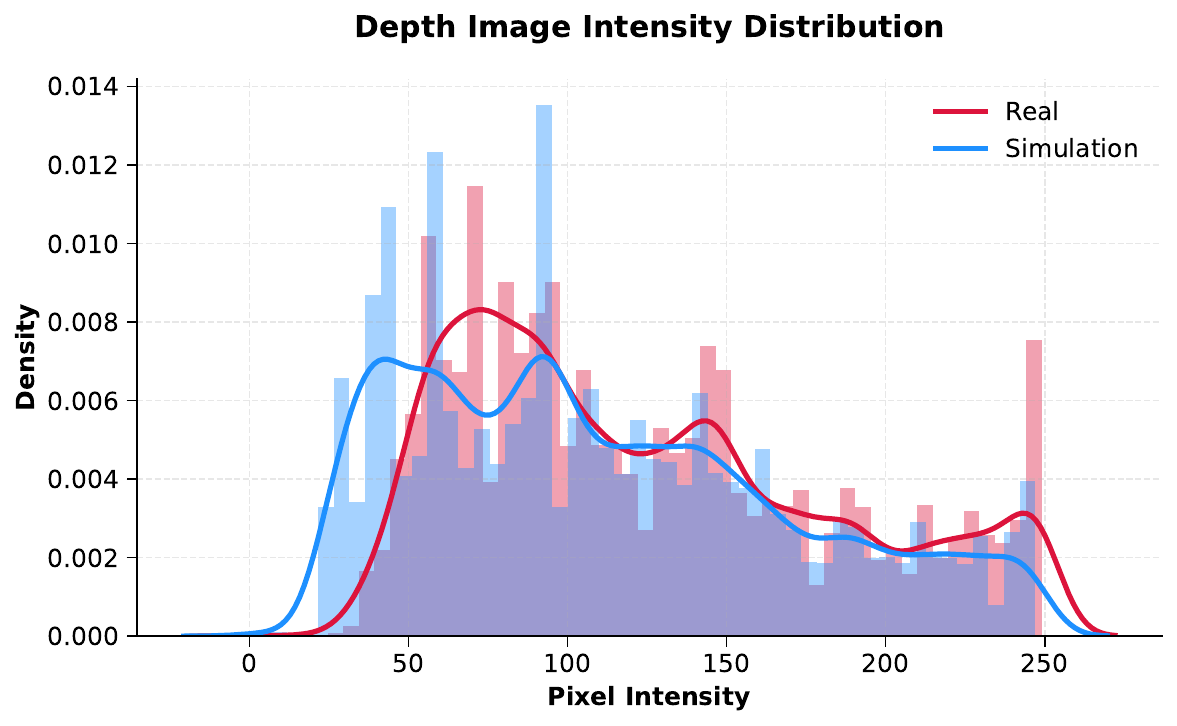}
  \vspace{-15pt}
  \caption{\textbf{Depth intensity distribution. }The horizontal axis represents the depth values, while the vertical axis indicates their corresponding proportions. This figure illustrates that the depth distributions derived from simulation and real-world images exhibit a notable resemblance in value patterns.}
  \label{fig:depth_dist}
\end{figure}
\subsection{Leveraging Depth Image as Visual Input}
Prior work~\cite{lumdextrah,zhuang2023robot,zhuanghumanoid,rudin2025parkour} has explored the use of depth images
for vision-based sim2real transfer. However, they often necessitate intricate and highly customized augmentation strategies to bridge the gap. In this work, we introduce a more versatile, manipulation-tailored egocentric depth-image augmentation method. Specifically, we clip depth values beyond a threshold distance $\mathbf{d}$~(set per task). For real depth images, missing depth values resulting from edge capture failures are filled in with the maximum depth. To emulate real-world edge noise and blur in simulation, we augment simulated depth images by adding Gaussian noise and Gaussian blur during training. Additionally, to mimic missing depth values, we randomly set 0.5\% of pixel values in simulation-rendered images to the maximum depth. To enrich the diversity of depth-noise distributions, we further employ the NYU Depth Dataset~\cite{silberman2012indoor} and adopt a mixup strategy that linearly blends the simulation-rendered depth image $\boldsymbol{o}_{sim}$ with a certain data depth map $\boldsymbol{o}_{dataset}$: ${\boldsymbol{\hat{o}}} = \alpha*\boldsymbol{o}_{sim} + (1 - \alpha)*\boldsymbol{o}_{dataset}$, where $\alpha$ is the coefficient. As illustrated in Figure~\ref{fig:compare_depth}, our augmentation  not only semantically aligns simulated renderings with real-world depth images, but also preserves crucial depth disparity cues essential for accurate visuomotor control. 

Furthermore, we visualize the distribution of depth values under similar frames in both simulation and real-world settings. As illustrated in Figure~\ref{fig:depth_dist}, our processing approach leads to a close alignment in value distributions between real-world and simulated depth images, indicating a reduced sim2real gap.

\subsection{DAgger Distillation Training}

In DAgger training, the state-based expert policy acts as the teacher to guide the learning of a visual student policy.  In contrast to prior approaches that distill to object masks or segmented images, \ourshort directly distills the state into raw visual observations of entire visual scenarios. This design obviates the need for explicit camera calibration and facilitates the acquisition of the robot's in-the-wild generalization capabilities. Furthermore, we introduce a series of auxiliary design choices aimed at enhancing both the asymptotic performance of DAgger training. 

\textbf{Model architecture: } The input observations are rendered at a resolution of $140\times 140$ pixels and subsequently stacked into sequences of 3 consecutive frames before being passed to the image encoder. Consistent with the design of Yuan et al.~\cite{yuanlearning,yuan2022pre}, we utilize the first two layers of a ResNet‑18 encoder~\cite{he2016deep} to more effectively capture fine‑grained visual details. Moreover, to ensure distributional consistency between training and evaluation phases, we replace all BatchNorm layers in the encoder with GroupNorm~\cite{wu2018group}.

\textbf{Trajectory rollout scheduler:} At the beginning of DAgger training, we adopt the expert policy to roll out trajectories for supervising the student. As training progresses, the reliance on expert rollouts is gradually reduced by annealing the probability $p$, while proportionally increasing the student policy's participation in rollouts. The pseudocode for the DAgger training procedure is presented in Algorithm~\ref{alg:training}, where $N_{\text{epochs}}$ denotes the total number of training epochs, $T$ represents the trajectory rollout intervals, and $\mathcal{D}$ refers to the replay buffer. At each interaction step with the environment,
the action $a$ is determined by the trajectory rollout scheduler, which probabilistically selects between the student actions $a_{\text{student}}$ and expert actions $a_{\text{expert}}$ based on a dynamically adjusted probability $p$. Here, we employ an exponential decay schedule to reduce the probability. 

\begin{algorithm}
\caption{DAgger Distillation Training}\label{alg:training}
\begin{algorithmic}

\State $\mathcal{D} \gets \emptyset$
\State $\text{trajectory rollout scheduler } \mathcal{S} \gets \text{Scheduler}(p_0)$
\For{$n \gets 0$ to $N_{\text{epochs}}$} 
    \For{$t \gets 0$ to  $T$} \Comment{Rollout Trajectory}
        \State $a_{\text{student}} \gets \pi_\text{student}(o_{\text{student}})$
        \State $a_{\text{expert}} \gets \pi_\text{expert}(o_{\text{expert}})$
        \State $a \gets \mathcal{S}$.\textproc{random\_choose} $(a_{\text{student}}, a_{\text{expert}}, p_t)$
        
        \State $\mathcal{D} \gets \mathcal{D} \cup \{(o_{\text{student}}, a_{\text{expert}})\}$
        \State $o_{\text{student}, t+1}, o_{\text{expert}, t+1} \gets$ \textproc{env.step}($a$)
        \State $p_{t+1} \gets \mathcal{S}$.\textproc{step}$()$ 
    \EndFor
    \State $\pi_\text{student} \gets \textproc{update}(\pi_\text{student}, \mathcal{D})$ \Comment{Policy Training}
\EndFor
\end{algorithmic}
\end{algorithm}

We jointly optimize the student policy using a combination of L1 and L2 action loss terms. Additionally, to mitigate the  overfitting to the low-dimensional states, we inject uniform noise into the proprioception states during training. More details and experimental results can be found in Appendix~\ref{sec:dagger} and Appendix~\ref{sec:dagger_train}. 

\begin{figure}[h]
  \centering
  \includegraphics[width=1.0\linewidth]{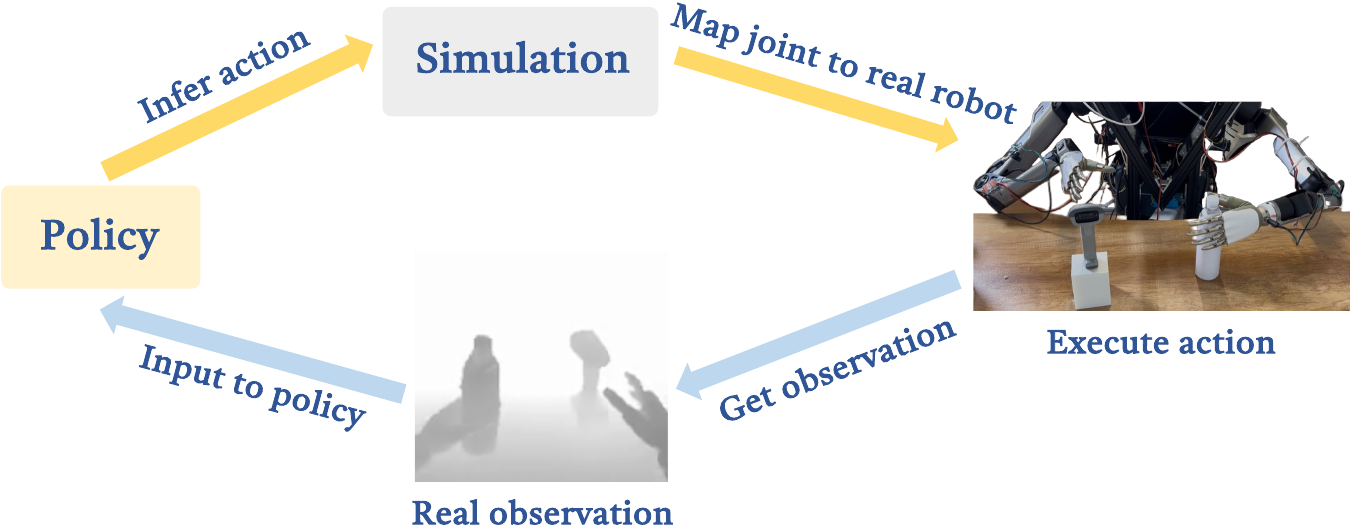}
  \vspace{-15pt}
  \caption{\textbf{Hybrid Sim2real Control.} We leverage real-world observations to infer actions and utilize simulation to compute the corresponding joint values, which are subsequently mapped onto the real robots. This hybrid strategy effectively mitigates the sim2real gap.}
  \label{fig:control}
\end{figure}
\subsection{Hybrid Sim2real Control}
Given the quasi-static nature of our tasks, we adopt a hybrid control strategy to mitigate the gap between simulation and real-world dynamics: real-world visual observations are used to infer the actual action, which is then applied to the simulation environment to perform a forward step. The updated joint positions of the simulated robot are subsequently transferred to the real robot for execution. Following this, the camera captures the image of the updated physical status of the real robot and incorporates the proprioception states as the input observation for the next inference cycle. The pseudocode of our control is provided in Algorithm~\ref{alg:control}.
By sharing the same Inverse Kinematics~(IK) method and dynamic parameters across simulation and the real world, this approach not only enables the policy to adapt its behavior based on real-world environmental variations but also effectively narrows the sim2real discrepancy. The pipeline is shown in Figure~\ref{fig:control}.

\begin{algorithm}
\caption{Hybrid Sim2real Control}\label{alg:control}
\begin{algorithmic}

\For{$t \gets 0$ to  $T$} \Comment{Real-world Policy Execution}
    \State $a_{t} \gets \pi_\text{policy}(o_{\text{real\_obs},t})$
     \State $o_{\text{sim}, t+1} \gets$ \textproc{env.step}($a_{t}$)
    \State $\mathcal{J}_{\text{all\_joints}} \gets$  \textproc{get\_all\_current\_joint\_qpos\_in\_sim}()
    \State $o_{\text{real\_obs},t+1} \gets$ \textproc{set\_all\_qpos\_in\_real}($\mathcal{J}_{\text{all\_joints}}$)
\EndFor
\end{algorithmic}
\end{algorithm}
\section{Navigation Methodology}
This section elaborates on how to equip the visuomotor control policy with the capacity to perform mobile manipulation tasks.

\subsection{ViNT Navigation Foundation Model}
To endow the trained visuomotor policy with navigation capabilities, \ourshort integrates an image-goal navigation foundation model~\cite{shah2023gnm,shah2023vint,sridhar2024nomad} that operates solely on RGB inputs and supports long-horizon, in-the-wild navigation. This framework allows for a seamless and low-cost fusion of manipulation and navigation modules, without necessitating additional fine-tuning of either component. We choose ViNT~\cite{shah2023vint} for achieving image-goal robotic navigation.  ViNT searches for the goal observations in the constructed topological map and computes a sequence of relative waypoints based on the current and goal observations, which are then translated into actions to control the low-level mobile controller. We deploy ViNT on our customized robotic system, operating at a frequency of 7.6 Hz. ViNT not only enables long-range, in-the-wild navigation but also demonstrates effective zero-shot generalization capability without necessitating model fine-tuning. 
\begin{figure*}[t]
  \centering
  \includegraphics[width=1.0\linewidth]{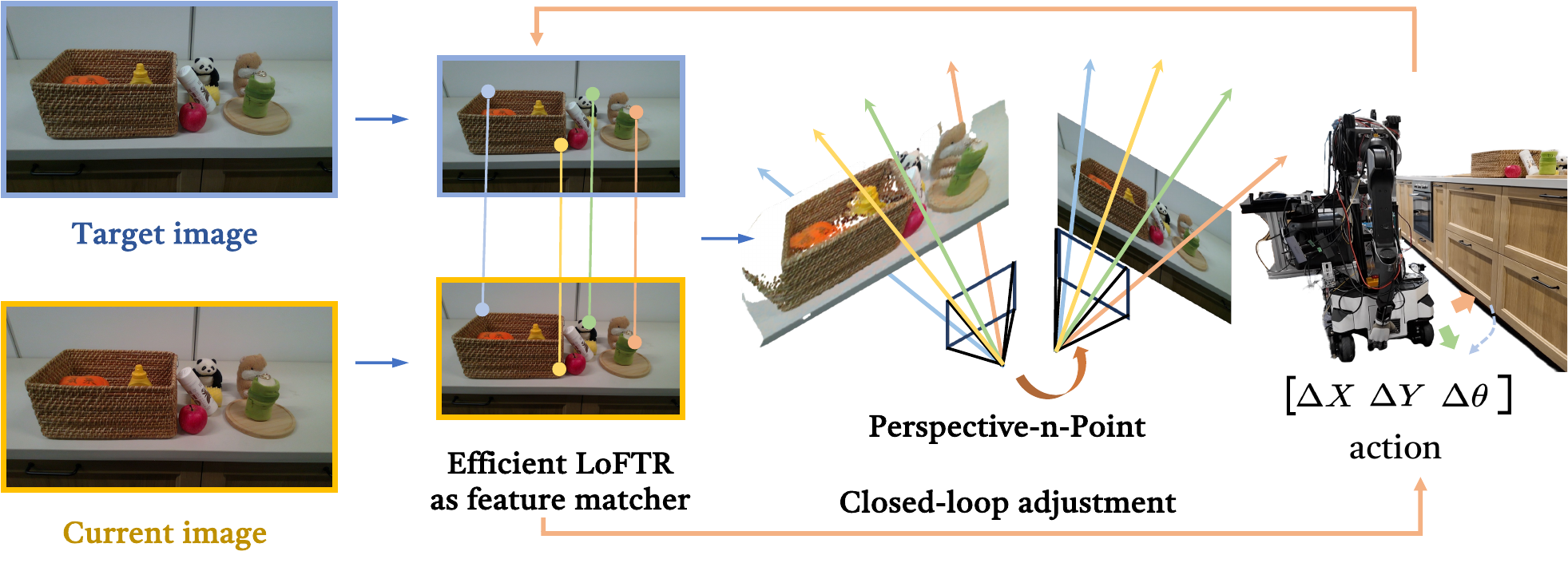}
  \vspace{-15pt}
  \caption{ \textbf{The pipeline of closed-loop PnP localization.} We first employ the Efficient LoFTR  to extract dense visual correspondence, followed by estimating the transformation between the current frame and the goal location via solving the PnP problem. Subsequently, we use PID controller to execute the action. This entire process is executed in a closed-loop manner and continues iteratively until the spatial discrepancy between the robot's current pose and the goal falls below a predefined threshold.}
  \label{fig:realtime-pnp}
\end{figure*}

\subsection{Closed-loop PnP Localization}
For our mobile manipulation tasks, moderate discrepancies between the robot's final pose and the target pose can lead to the manipulation policy failing to finish the task. However, ViNT does not guarantee termination within a sufficiently tight error bound. To address this, we introduce a local refinement step after ViNT completes navigation: a closed-loop Perspective-n-Point~(PnP) localization algorithm is employed to adjust the robot pose, ensuring closer alignment with the target pose.

As shown in Figure~\ref{fig:realtime-pnp}, we first utilize the neural feature matching module Efficient LoFTR~\cite{wang2024efficient} to detect the correspondence between the current robot captured image $I_{c}$ and the goal image $I_{g}$. Next, the detected features are lifted to 3D space with respect to the robot's current coordinate frame by leveraging the camera intrinsic matrix $\mathbf{K}$ and the depth map $\mathbf{D_{c}}$. This yields $\mathbf{X_a}$, the 3D coordinates of the matched features in the current camera coordinate frame. According to Equation~\ref{eq: pnp}, we then leverage the RANSAC PnP~\cite{zuliani2009ransac} and refine PnP algorithm~\cite{madsen2004methods,eade2013gauss} to compute the relative rotation $\mathbf{R} \in \mathbb{R}^{3 \times 3}$ and translation $\mathbf{t} \in \mathbb{R}^{3}$ between the robot’s current viewpoint and the goal pose that can minimize the reprojection error $e=\left\|\tilde{\mathbf{P}}_{\mathbf{g}}-\mathbf{P}_{\mathbf{g}}\right\|_2^2 $: 

\begin{equation}
\label{eq: pnp}
\left[\begin{array}{c}
\tilde{\mathbf{P}}_{\mathbf{g}} \\
1
\end{array}\right]=\mathbf{K}[\mathbf{R} \mid \mathbf{t}]\left[\begin{array}{c}
\mathbf{X_a} \\
1
\end{array}\right] ,   
\end{equation}

where $\mathbf{P_{g}}$ denotes the pixel positions of the matched features at the goal image,  $\tilde{\mathbf{P}}_{\mathbf{g}}$ represents the reprojected positions. 
Adopting Efficient LoFTR can yield an inference rate of 16.7 Hz. By leveraging real-time feedback from PnP as the robot incrementally converges toward the target pose, we are able to iteratively refine the pose estimation, thus attaining more accurate visual correspondence. 

After getting the target pose calculated by our closed-loop PnP localization algorithm, we utilize a Proportional-Integral-Derivative~(PID) controller~\cite{willis1999proportional} to adjust the pose of our robot. The input of the controller is the instantaneous position and orientation error between the robot's desired state and its actual state. Since the mobile base exhibits limited control accuracy, we define separate PID controllers for reducing errors individually along the x and y directions, as well as for the yaw angle. Based on this multi-dimensional error input, the PID controller computes and outputs corresponding \textit{planar velocity commands} designed to minimize the error. These commands consist of velocities along the corresponding three directions. The velocity is computed as follows:

\begin{align}
v_j &= K_{p,j} e_j + K_{i,j} \int_{0}^{t} e_j(\tau)  d\tau + K_{d,j} \frac{de_j}{dt} 
\end{align}

where $j$ denotes a specific axis, $v_{j}$ is the velocity, $e_{j}$ represents the error term, and $K_{p,j}$, $K_{i,j}$, $K_{d,j}$ correspond to the proportional, integral, and derivative gain coefficients, respectively.

To address the characteristics of our omnidirectional chassis, which incurs additional displacement during wheel reorientation, we implement a sequential adjustment strategy, which prioritizes error correction in the order of x-direction, y-direction, yaw orientation. This staged compensation mitigates the coupling effects introduced by wheel realignment. 

\section{Experiments}
In this section, we perform an extensive series of experiments aimed at evaluating the capabilities of \ourshort across various aspects, including navigation and manipulation. Specifically, our primary experiments are designed to: (1) verify the efficacy of \ourshort in efficiently and robustly transforming diverse human motion data into robot-plausible behaviors; (2) exhibit the effectiveness of our method in sim2real transfer; (3) quantify the accuracy and reliability of our navigation localization approach; (4) demonstrate the effectiveness of \ourshort in mobile manipulation. 

\subsection{Sample Efficiency of \ourshort}
\begin{figure*}[t]
  \centering
  \includegraphics[width=1.0\linewidth]{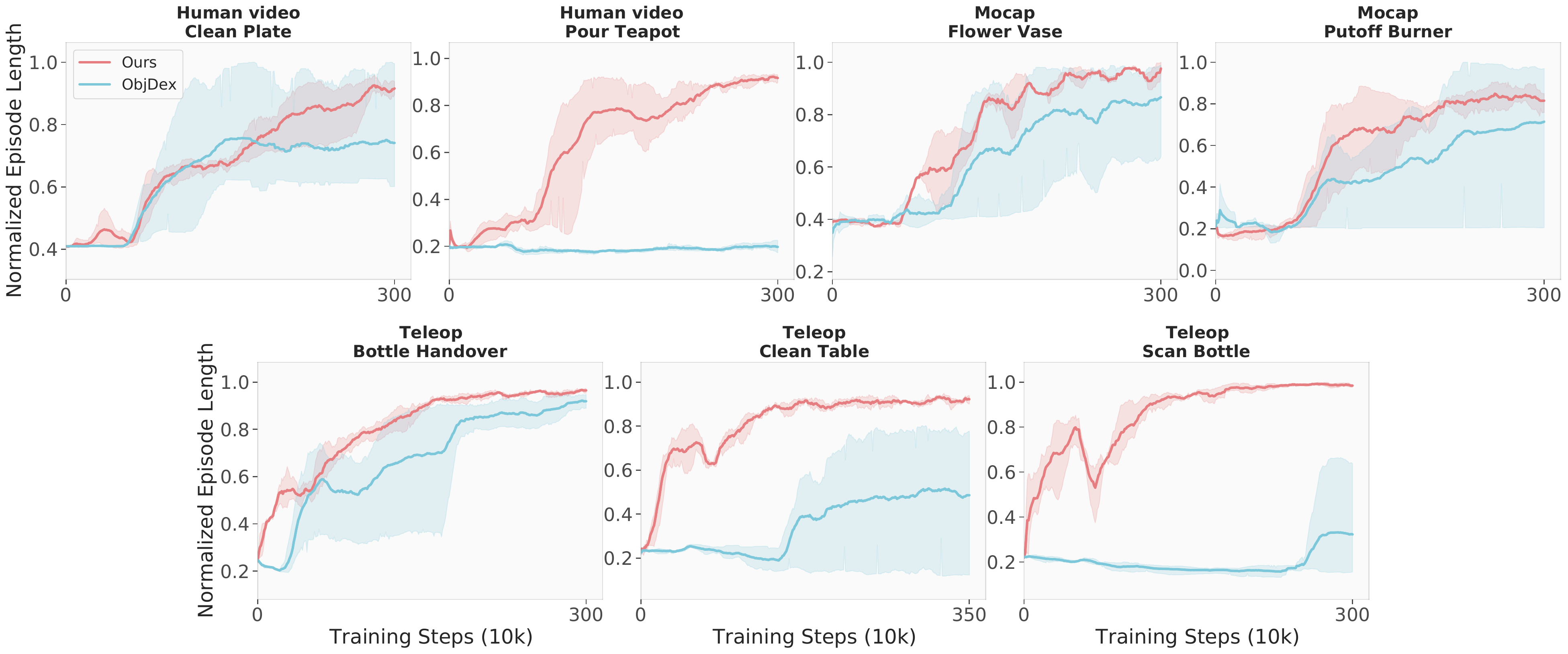}
  \caption{ \textbf{The training curve of \ourshort.} The horizontal axis denotes the training steps, while the vertical axis represents the normalized task length successfully accomplished by the policy. \textit{Teleop} refers to one-shot human motion teleoperation in simulation, \textit{Human video} denotes trajectories extracted from video data, and \textit{Mocap} corresponds to motion derived from mocap datasets. \ourshort not only demonstrates the capability to accomplish diverse manipulation tasks originating from various forms of human motion, but also exhibits superior sample efficiency throughout training. All results are evaluated across 3 seeds. }
  \label{fig:train_curve}
\end{figure*}
\begin{figure*}[t]
  \centering
  \includegraphics[width=1.0\linewidth]{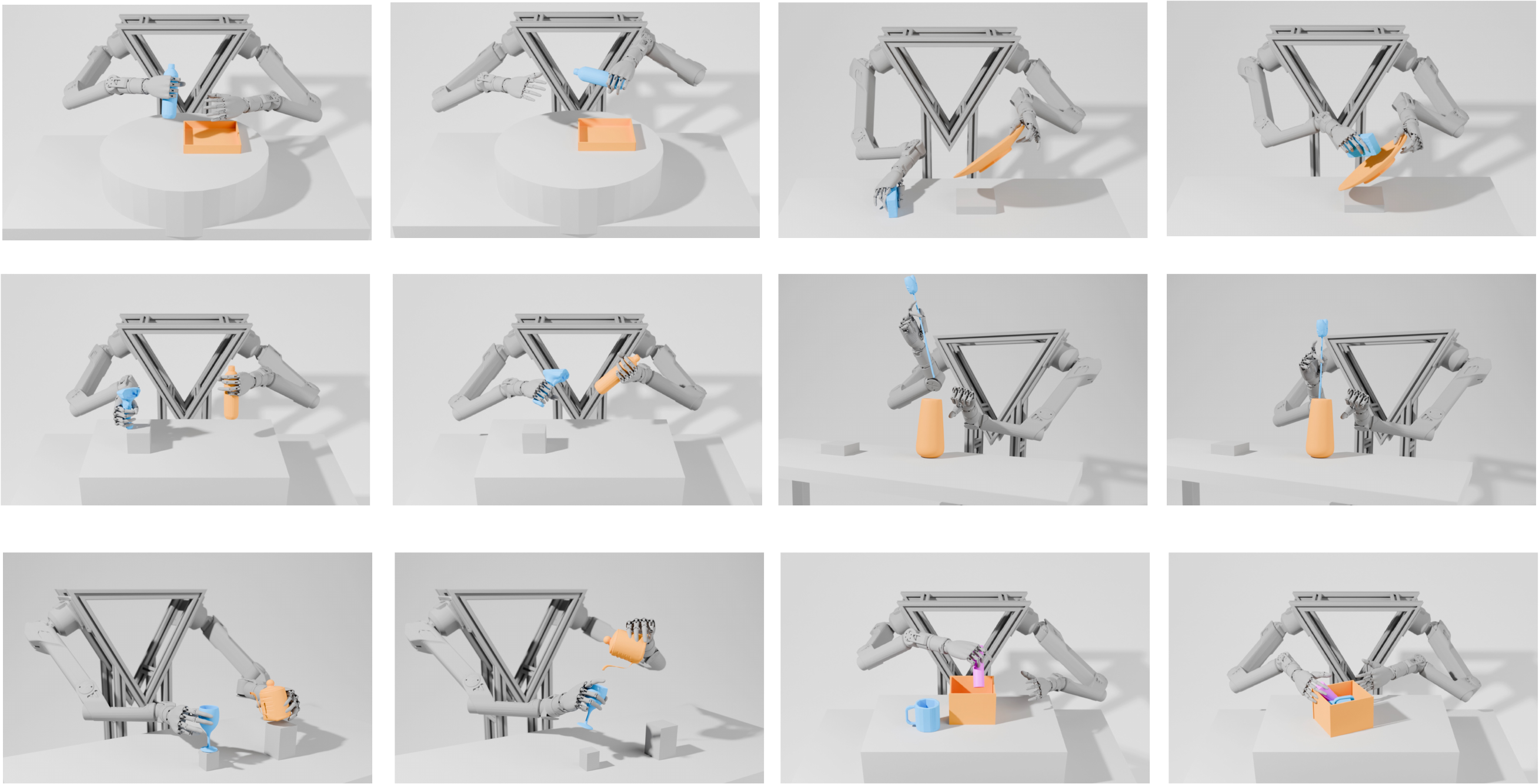}
  \caption{ \textbf{Simulation training visualization.} We visualize the majority of the training tasks. Leveraging a single reference trajectory in conjunction with a general reward design, \ourshort can convert diverse human motion sources into robot feasible behaviors via RL training.}
  \label{fig:sim_vis}
\end{figure*}

We evaluate the training sample efficiency of \ourshort across seven tasks. The visualizations of simulation tasks are shown in Figure~\ref{fig:sim_vis}. For each task, the source of the one-shot human motion demonstration is indicated in the title of each sub-figure in Figure~\ref{fig:train_curve}. The vertical axis in the figure represents the proportion of the trajectory length successfully executed by the current policy relative to the total length of the trajectory. As demonstrated in Figure~\ref{fig:train_curve}, regardless of the origin of the human motion data, \ourshort reliably succeeds in converting human hand and arm actions into generalizable robot-executable behaviors.

Additionally, we compare training performances with ObjDex~\cite{chenobject}. ObjDex defines its reward based on the tracking of the object's joint movement, translations, and orientations. We re-implement this reward formulation within our own algorithmic framework. Figure~\ref{fig:train_curve} indicates that \ourshort exhibits superior performance relative to ObjDex across all tasks. In tasks such as \textit{Bottle Handover}, \textit{Flower Vase}, and \textit{Putoff Burner}, where interactions involve only a single object, ObjDex is able to complete the tasks; however, \ourshort can achieve higher sample efficiency during training. Furthermore, in more intricate tasks involving multi-object interactions, ObjDex consistently fails, irrespective of the type of human motion data provided. Owing to our object-centric distance chain, \ourshort is capable of robustly acquiring diverse manipulation skills even in long‑horizon, multi-object environments. Moreover, \ourshort demonstrates high sample efficiency and successfully learns policies in 3M training steps.

\begin{figure}[h]
  \centering
  \includegraphics[width=1.0\linewidth]{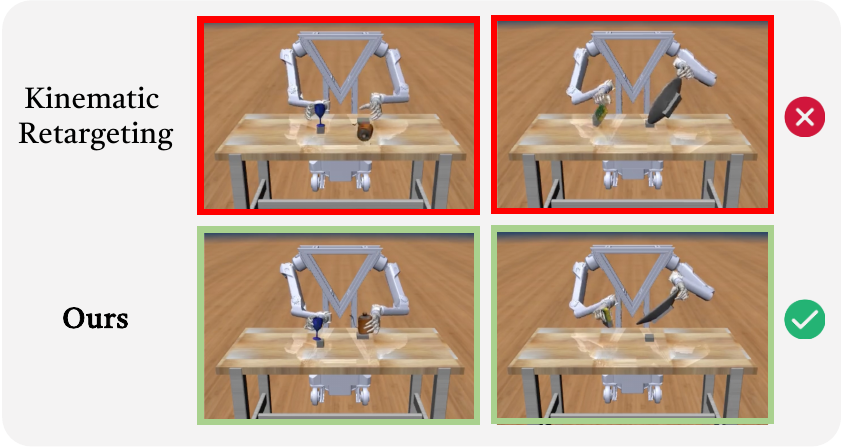}
  \caption{ \textbf{The comparison of kinematic retargeting and \ourshort.} The raw trajectories extracted from human videos and mocap data are insufficient to complete the task through mere kinematic retargeting. On the other hand, \ourshort not only learns to follow these reference trajectories but also masters the nuances of object interaction.}
  \label{fig:kinematics}
\end{figure}

\subsection{Comparison with Non-learning Approach}
In this section, we highlight the central role of reinforcement learning in equipping robots with adaptive policies and delineate the advantages it affords over non-learning approaches.

\begin{table}[h]
\centering
\caption{Comparison of \ourshort and kinematic retargeting.}
\label{table: compare_retarget}
\resizebox{0.5\textwidth}{!}{
\begin{tabular}{c|cc}
\toprule

\setlength{\tabcolsep}{10pt}
\begin{tabular}[c]{@{}c@{}} \multicolumn{1}{c}{\multirow{1}{*}{Tasks $\backslash$ Method}} \end{tabular}   & \multicolumn{1}{c}{\ourshort} & \multicolumn{1}{c}{kinematic retargeting} \\

\midrule

Video tasks & \ddbf{78.1}{15.0} & $0$ \\

Mocap tasks &	\ddbf{88.9}{6.7} & $0$ \\

\bottomrule
\end{tabular}
}
\end{table}

For human motion trajectories derived from mocap data and videos, we compare \ourshort with kinematic retargeting methods. Table~\ref{table: compare_retarget} indicates that while kinematic retargeting can map human hand motions to robotic counterparts, it fails to capture essential aspects such as object interactions and contact information. Moreover, the retargeting approaches cannot guarantee optimality. The snapshots of two approaches are visualized in \ref{table: compare_replay}. Hence, RL emerges as a crucial approach for refining robotic behaviors. It shapes the robot policies toward human-like motions and establishes physically plausible, context-appropriate object interactions.

\begin{table}[h]
\centering
\caption{Comparison of \ourshort and replay edited trajectories.}
\label{table: compare_replay}
\resizebox{0.4\textwidth}{!}{
\begin{tabular}{c|cc}
\toprule

\begin{tabular}[c]{@{}c@{}} \multicolumn{1}{c}{\multirow{1}{*}{Tasks $\backslash$ Method}} \end{tabular}   & \multicolumn{1}{c}{\ourshort} & \multicolumn{1}{c}{replay} \\

\midrule

Bottle Handover & \ddbf{91.9}{1.9} & \dd{52.2}{5.1}  \\

Place Drawer &	\ddbf{72.2}{5.1} & \dd{49.9}{5.8}  \\

\bottomrule
\end{tabular}
}
\end{table}

Regarding teleoperation data, we compare \ourshort with direct replay of edited trajectories. We evaluate both methods on two distinct tasks: a long-horizon task involving articulated object manipulation~(\textit{placedrawer}) and a contact-rich task~(\textit{handover}). For each task, the object's pose is randomized on the table, and each method is evaluated over 30 episodes. As demonstrated in Table~\ref{table: compare_replay}, due to the robot's dynamic configuration, directly executing edited actions does not guarantee task success when object positions change. In contrast, \ourshort, through RL training, learns residual actions that can adaptively adjust movements and enhance execution success rates. These results also highlight that \ourshort leveraging RL can effectively mitigate dynamic inconsistencies. 
\begin{figure}[h]
  \centering
  \includegraphics[width=1.0\linewidth]{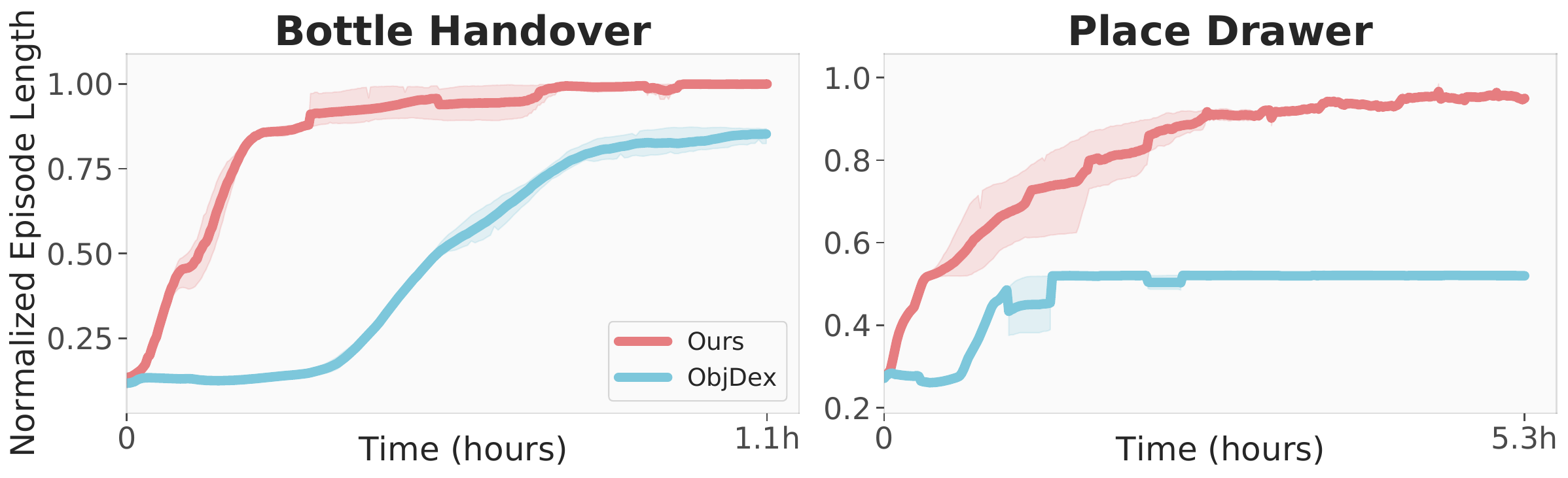}
  \caption{ \textbf{The wall-time training efficiency.} \ourshort also enjoys  high wall-time efficiency under parallel training. All results are evaluated across 3 seeds.}
  \label{fig:walltime}
\end{figure}
\subsection{Training wall-time}
We also leverage the MJX GPU parallel simulation~\cite{MJX} to instantiate the tasks and adopt the PPO algorithm~\cite{schulman2017proximal} for policy training. We adopt reward terms in line with DrM. As illustrated in Figure~\ref{fig:walltime}, \ourshort benefits from reduced wall-clock training time, and the reward formulation demonstrates cross-algorithm generality. Meanwhile, compared to the baseline method, \ourshort also attains higher sample efficiency and stronger asymptotic performance under PPO training.

\begin{figure*}[t]
  \centering
  \includegraphics[width=1.0\linewidth]{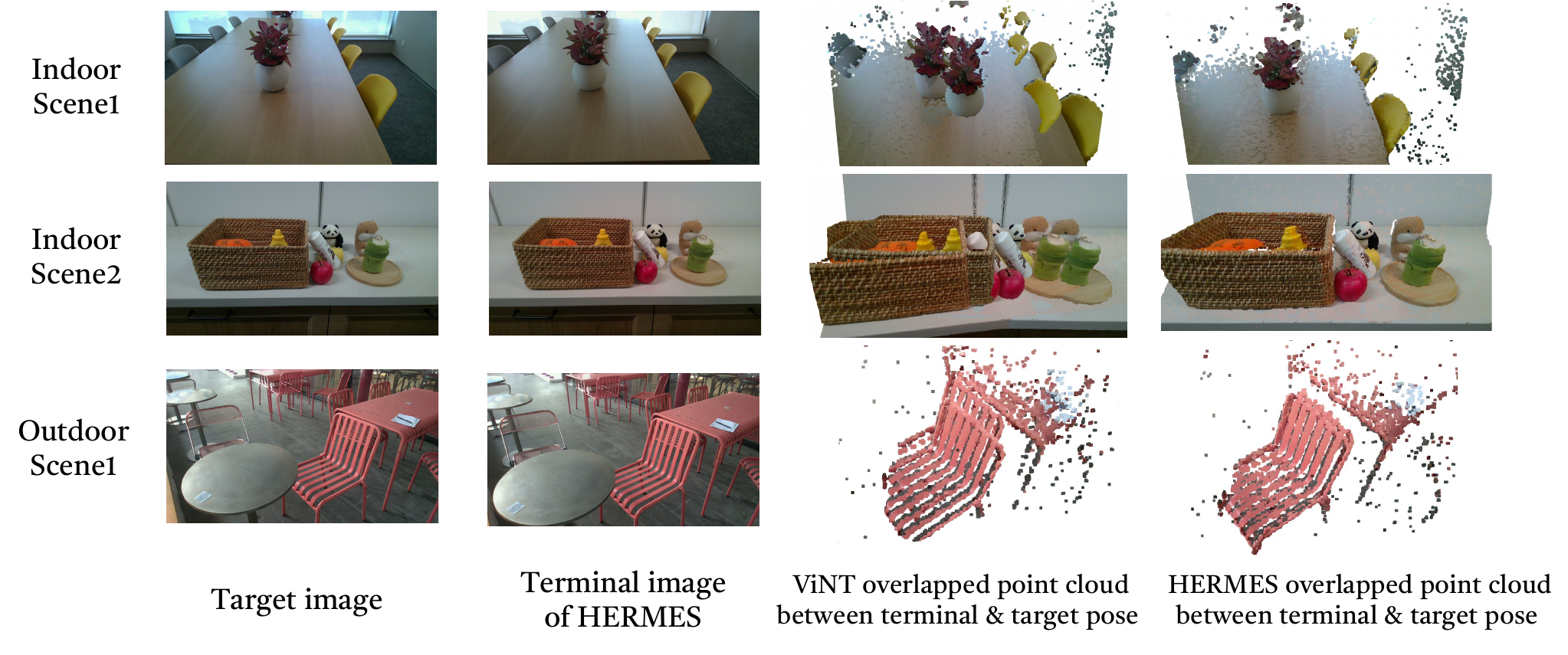}

  \caption{ \textbf{The visualization of navigation results.} The left two columns depict a comparison between the target image and the terminal image achieved by our method. The right two columns present the point clouds captured at the end of navigation by ViNT and \ourshort, compared against the point cloud of the target position. This figure illustrates that ViNT exhibits a noticeable mismatch between the captured and target point clouds at the end of navigation, whereas \ourshort achieves a close alignment, which demonstrates the high localization accuracy of our approach.}
  \label{fig:nav_result}
\end{figure*}

\subsection{Real-world Manipulation Evaluation}

\begin{table*}[t] \caption{Real-world manipulation  evaluation results. Across $6$ real-world bimanual dexterous manipulation tasks, \ourshort obtains $\mathbf{+54.5\%}$  performance gains on average. }
\label{table1: generalization}
\centering
\renewcommand\tabcolsep{4.0pt}
\renewcommand{\arraystretch}{1.2}
\begin{normalsize}
\resizebox{0.9\textwidth}{!}{
\begin{tabular}{c|cccccc|c}
\cline{2-6}
\toprule[0.3mm]
\begin{tabular}[c]{@{}c@{}}  \multicolumn{1}{c}{\multirow{1}{*}{Method  $\backslash$ Tasks}} \end{tabular} & Bottle Handover & Clean Table & Scan Bottle & Putoff Burner & Clean Plate & Pour Teapot &  \textbf{Average} \\
\hline 
\begin{tabular}[c]{@{}c@{}} \ourshort  \end{tabular} & \best{66.7}  & \best{60.0} & \best{73.3}  & \best{66.7} & \best{66.7} & \best{73.3}   & \textcolor{lyydeepred}{\ddbf{67.8}{5.0}} \\
\begin{tabular}[c]{@{}c@{}} raw depth \end{tabular}  & $6.7$ & $0.0$  &  $0.0$  & $20.0$   &   $13.3$    & $40.0$   &   \dd{13.3}{15.2}   \\
\hline
\end{tabular}
}

\end{normalsize}
\label{table:sim-results}
\vspace{-10pt}
\end{table*}

After conducting DAgger training, we subsequently transfer the trained visual student policy to the real world in a zero-shot manner for most tasks. It should be noted that for the tasks \textit{pour teapot} and \textit{putoff burner}, the presence of substantial noise in the trajectory or transparent objects leads to excessively jittering motions, along with discrepancies between simulated and real-world object shapes. Consequently,  we additionally fine-tune the policy using 5 extra real-world trajectories collected via policy rollouts. 

Table~\ref{table:sim-results} presents the generalization performance of the policy evaluated across different object placements and poses, with each task assessed over 15 trials. As the baseline, we replace our depth-processing with raw depth inputs in the sim2real pipeline. \ourshort not only successfully achieves zero-shot transfer for diverse long-horizon or contact-rich bimanual dexterous manipulation tasks, but also surpasses the baseline by $\mathbf{+54.5}$\% in success rate. These experimental results substantiate \ourshort's capability to effectively bridge both visual and dynamic gaps, enabling successful sim2real transfer and demonstrating intricate manipulation skills. Moreover, for the two tasks involving fine-tuning with real-world rollout trajectories, owing to the reduced visual discrepancy achieved by \ourshort, the trained policy exhibits enhanced generalization capabilities compared to the raw depth baseline.

\subsection{The Effectiveness of Closed-loop PnP}

In terms of the navigation experiments, we first evaluate the localization errors of the ViNT model augmented with our proposed closed-loop PnP localization algorithm. We conduct experiments in two indoor and one outdoor scenarios, including two long-horizon navigation tasks. Table~\ref{table:nav_error} reports the computed localization errors in both translational distance and orientation. These errors are derived by solving the PnP problem and subsequently calculating the relative pose differences between the terminal pose and the target pose within a shared reference frame. As shown in Table~\ref{table:nav_error}, incorporating our proposed approach significantly reduces localization errors, whereas ViNT suffers from substantial instability in localization accuracy. Additionally, we visualize the RGB images and corresponding point clouds of the stopping positions for both baselines and \ourshort. Figure~\ref{fig:nav_result} demonstrates that \ourshort not only achieves semantic alignment of RGB images but also accurately matches the point clouds with the target position due to improved localization precision. The experimental results presented in Figure~\ref{fig:nav_result} and Table~\ref{table:nav_error} highlight that current general navigation models, despite their ability to broadly match the goal image, exhibit large localization discrepancies, rendering them unsuitable for downstream visuomotor manipulation tasks. In contrast, through employing our simple yet robust closed-loop PnP localization algorithm, \ourshort not only successfully addresses the low precision of the mobile robot base, but also achieves small localization errors.

\begin{table}[ht]
\centering
\caption{The results of navigation localization error.}
\label{table:nav_error}
\resizebox{0.48\textwidth}{!}{
\begin{tabular}{l|cc|cc}
\toprule
\begin{tabular}[c]{@{}c@{}} \multicolumn{1}{c}{\multirow{2}{*}{Method  $\backslash$ Tasks}} \end{tabular} & \multicolumn{2}{c|}{\textbf{\ourshort}} & \multicolumn{2}{c}{\textbf{ViNT}}  \\
& \small dist error~(cm) & \small ori error~($^\circ$)  &\small dist error~(cm) &\small ori error~($^\circ$)  \\
\midrule
indoor scene1 & \blueddbf{2.4}{0.5} & \blueddbf{1.79}{1.12} & \dd{18}{11.3} & \dd{2.57}{1.98}  \\
indoor scene2 & \blueddbf{1.3}{0.5} & \blueddbf{0.57}{0.57}  & \dd{7.3}{3.1} & \dd{3.66}{1.91} \\
outdoor scene1 & \blueddbf{3.2}{1.8} & \blueddbf{1.67}{1.84}  & \dd{12.9}{7.4} & \blueddbf{1.63}{1.31} \\

\bottomrule
\end{tabular}}

\end{table}

\subsection{The Localization Ability of Closed-loop PnP in the Textureless Scenario}
\begin{figure}[h]
  \centering
  \includegraphics[width=0.9\linewidth]{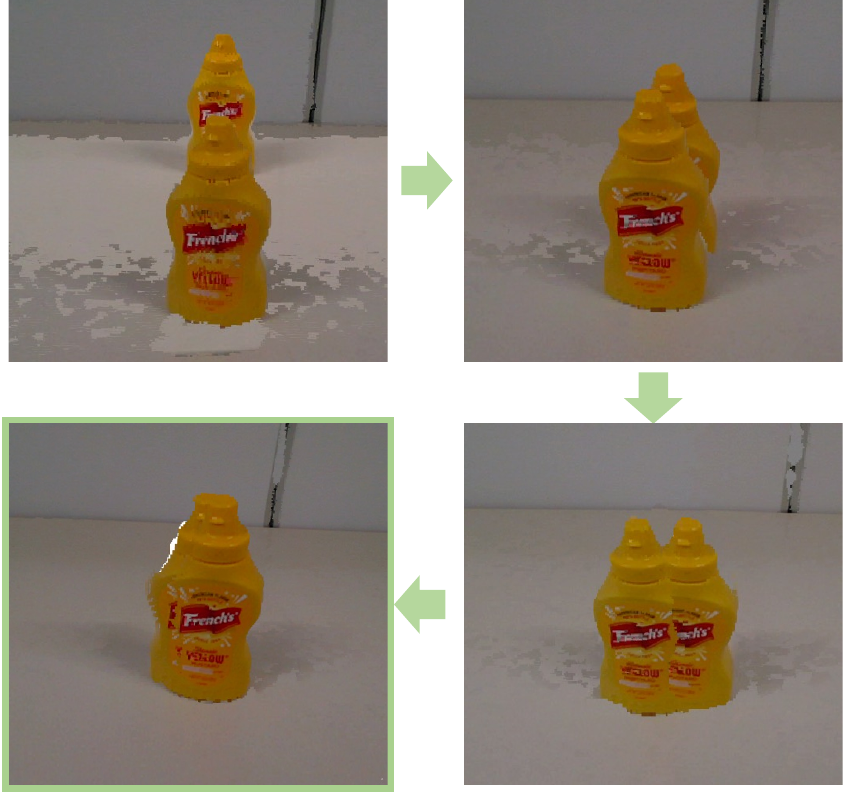}
  \caption{ \textbf{The localization ability of \ourshort in textureless scenarios.} Even in environments with sparse visual features, \ourshort remains capable of executing fine-grained positional adjustments and achieving precise localization through the closed-loop PnP mechanism.}
  \label{fig:plain}
\end{figure}
We also compare the localization performance of RTAB-MAP~\cite{Labb__2018}, a popular RGBD-based visual SLAM approach, against \ourshort in the textureless scenario. Previous studies have shown that RGBD-based SLAM methods struggle to accurately localize in textureless environments since it is hard to track visual features in these scenes~\cite{Labb__2018,Campos_2021,s19102251}. As shown in Table~\ref{table: nav_textureless}, RTAB-MAP fails to accomplish the task while our approach consistently achieves precise localization even under such challenging conditions. The point clouds of iterative refinement processes of closed-loop PnP are shown in Figure~\ref{fig:plain}.

\begin{table}[h]
\centering
\caption{Comparison of \ourshort and RTAB-MAP in the textureless scenario.}
\label{table: nav_textureless}
\resizebox{0.45\textwidth}{!}{
\begin{tabular}{c|cc}
\toprule

\begin{tabular}[c]{@{}c@{}} \multicolumn{1}{c}{\multirow{1}{*}{Method $\backslash$ Error}} \end{tabular}   & \multicolumn{1}{c}{dist error~(cm)} & \multicolumn{1}{c}{ori error~($^\circ$)} \\

\midrule

RTAB-MAP & - & -  \\

\ourshort &	\ddbf{1.26}{1.25} & \ddbf{2.06}{1.50}  \\

\hline
\end{tabular}
}
\end{table}

\begin{figure}[h]
  \centering
  \includegraphics[width=1.0\linewidth]{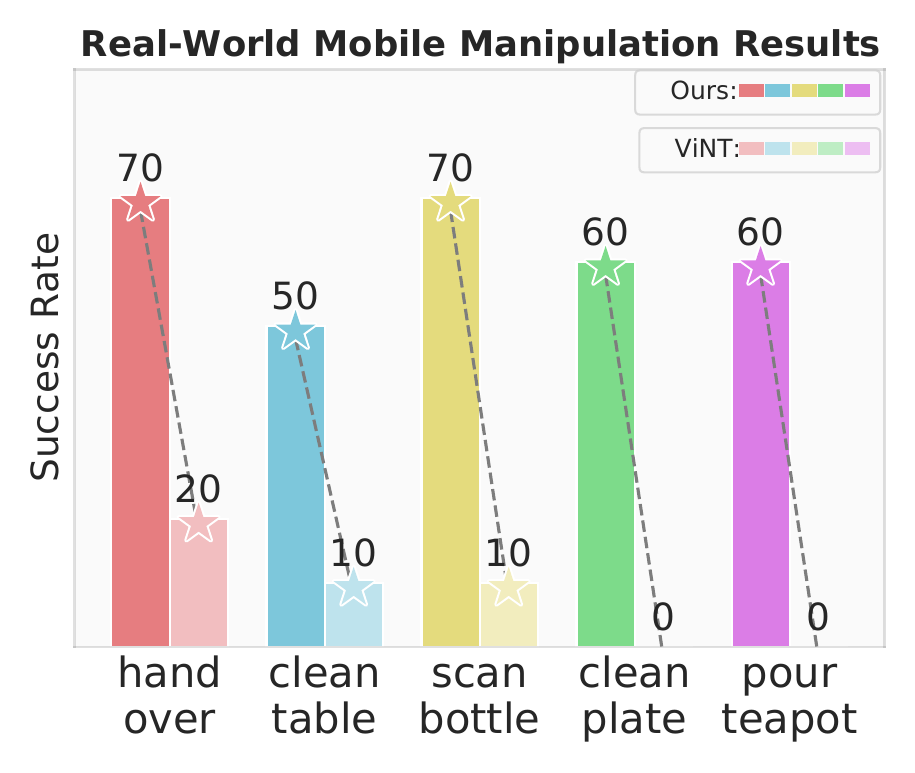}
  \vspace{-15pt}
  \caption{ \textbf{Real-world mobile manipulation results.} Dark-colored bars correspond to \ourshort, whereas the light-colored bars correspond to only using ViNT. \ourshort is capable of performing a wide array of complex mobile bimanual dexterous manipulation tasks. In contrast, when relying solely on ViNT for localization, the trained manipulation policy fails to complete the tasks. } 
  \label{fig:mani_bar}
\end{figure}
\subsection{Mobile Manipulation Evaluation}
To evaluate the mobile manipulation ability of \ourshort, we integrate the entire pipeline across all tasks. Each trained policy is tested over 10 runs. As illustrated in Figure~\ref{fig:mani_bar}, \ourshort demonstrates strong real‑world navigation, precise localization, and dexterous manipulation capabilities. We also apply the identical manipulation policy equipped with ViNT as a baseline. Figure~\ref{fig:mani_bar} reveals that, without closed-loop PnP localization, the policy cannot generalize or successfully complete tasks when faced with significant positional and rotational shifts. Conversely, \ourshort achieves a notable $\mathbf{+54.0}$\% improvement in manipulation success rate compared to pure ViNT. These findings underscore that closed-loop PnP localization is the essential bridge linking navigation and manipulation, enabling both modules to synergize for enhanced performance.
\section{Conclusion}
In this work, we introduce \ourshort, a novel framework addressing critical challenges in bimanual dexterous robotic manipulation by effectively leveraging diverse human motion data sources and robust sim2real methodologies. Through the integration of a hybrid control scheme alongside a generalized DAgger-based distillation framework, \ourshort facilitates effective sim2real transfer with high success rates. Additionally, we present a navigation localization method employing closed-loop PnP refinement, crucial for bridging the gap between navigation accuracy and manipulation precision. The comprehensive experimental results validate that \ourshort not only achieves superior performance in complex manipulation tasks but also demonstrates exceptional adaptability to diverse real-world scenarios, thus providing a solid foundation for future advancements in mobile robotic manipulation.

\section{limitations and future work}
 Although we have demonstrated the effectiveness of \ourshort, several limitations remain. First, our tasks are quasi-static, for which the proposed hybrid sim2real control scheme is well suited; however, for highly dynamic, velocity-dependent tasks, complicated system identification is still required for sim2real transfer. Second, to obtain favorable robot behaviors, we still manually tune physics collision parameters and approximate objects with primitive geometric shapes. Additionally, assembly and calibration mismatches between simulation and hardware persist; while closed-loop control can mitigate these discrepancies, they nevertheless reduce overall success rates. In future work, we will attempt to deploy our algorithms on more robust and precisely engineered hardware systems and to lower the cost of simulation setup, improving reliability and reducing manual effort.
\section{Acknowledgment}
We gratefully acknowledge OYMotion for providing the dexterous hand hardware that supported this work, and we thank Sizhe Yang, Zixuan Liu and Zhengmao He for their insightful comments on the manuscript. This work is also supported by dushi program.

\bibliographystyle{IEEEtran}
\bibliography{IEEEabrv,./references}

\newpage
\appendix
\subsection{State-based RL Expert Training Details}
\label{sec:state-rl}
\textbf{Model Architecture:} Since DrM~\cite{xudrm} is inherently designed as a visual reinforcement learning algorithm, we replace its visual encoder with a three-layer multilayer perceptron (MLP), employing ELU as activation functions. Meanwhile, the network architectures and hyperparameters of the Actor, Critic, and V-network remain unchanged.

\textbf{Training Hyperprameters:} We employ the identical set of reinforcement learning hyperparameters as those in the original visual RL implementation, without conducting any further hyperparameter tuning specifically for DrM in our experiments. For PPO, we employ the implementation available in MuJoCo Playground~\cite{mujoco_playground_2025}, which is built upon the RSL-RL framework~\cite{rudin2022learning}.

\textbf{Reward Hyperparameters:} The reward hyperparameters are listed in Table~\ref{table:reward_hype}. In addition to the proposed reward terms, contact information is incorporated during training on teleoped-motion tasks to tune the distance-chain reward coefficients.
\begin{table}[h]
    \centering
    \caption{Reward Hyperparameters}
    \begin{tabular}{c|c}
    \hline
    Description & Value \\
    \hline
        contact predefined threshold $N_{\text{num}}$ & $2$ \\
        object position reward coefficient $k_{1}$ & $1$ \\
        object orientation reward coefficient $k_{2}$ & $1$ \\
        coefficient of penalty & $1e-3$ \\

    \end{tabular}
    \label{table:reward_hype}
\end{table}

\textbf{Observations of RL training:} The dimensions of the observation for training expert state-based policy are provided in Table~\ref{table:obs_all}. 
\begin{table}[h]
    \centering
    \caption{Observation Descriptions}
    \begin{tabular}{c|c}
    \hline
    Description & Dimension \\
    \hline
        arm joint position  & $12$ \\
        hand joint position & $36$ \\
        arm joint velocity & $12$ \\
        hand joint velocity & $36$ \\
        hand position & $6$ \\
        hand quaternion & $8$ \\
        object-centric distance chain & $72$ \\
        contact information & $24$ \\
        actuator value & $24$ \\
        time & $1$ \\
        distance with the reference  & $10$ \\
        hand position in reference & $6$ \\
        hand quaternion in reference & $8$ \\
        object position in reference & $3$ \\
        object quaternion in reference & $4$ \\

    \end{tabular}
    \label{table:obs_all}
\end{table}

\textbf{Action space:} Actions are parameterized as a six-dimensional end-effector pose: the first three are translation, and the last three are Euler angles. For residual actions,  the translation range is $[-0.01, 0.01]$ m and the orientation range is $[-0.04, 0.04]$ rad. Regarding the actions in trajectories, teleoperated motions are scaled to $[-0.2, 0.2]$ m and $[-0.4, 0.4]$ rad, whereas motions derived from video and mocap are scaled to $[-1, 1]$ m and $[-1, 1]$ rad. 

\subsection{Human Motion Data Pre-processing}
We apply downsampling to the collected human motion data in order to reduce episode length and redundant steps, thereby facilitating more efficient policy learning of task skills. For human video data, particularly when involving symmetric objects, we observe that the FoundationPose~\cite{wen2024foundationpose} estimator may introduce rotational ambiguities during pose prediction. To address this, we preprocess the object pose trajectories to mitigate pose discrepancies induced by object symmetry.

\subsection{Task Description}
We design a suite of tasks specifically tailored for bimanual dexterous hand manipulation. Below, we provide a detailed description of each individual task:

\textit{Bottle Handover}: The robot is required to execute a hand-to-hand transfer, passing a bottle from the right hand to the left, followed by accurately placing it into a designated container.

\textit{Clean Table}: The task entails sequentially picking up both a bottle and a cup from the tabletop and placing them into a box.  

\textit{Scan Bottle}: The robot must first grasp a bottle from the table, subsequently pick up a scanning device, and then complete a bottle-scanning action. 

\textit{Place Drawer}: This task requires the robot to open a drawer, place multiple objects from the tabletop into the drawer sequentially, and finally close the drawer. 

\textit{Pour Teapot}: The robot is required to perform a pouring action by transferring liquid from a container into a teapot. 

\textit{Clean Plate}: The robot must grasp both a plate and a dishcloth, and then execute a wiping motion over the plate’s surface. 

\textit{Putoff Burner}: This task involves picking up a burner cap and performing a two-step extinguishing sequence to put out an alcohol burner

\textit{Flower Vase}: The robot is required to insert a flower into a vase.

\subsection{DAgger Training Details}
\label{sec:dagger}
We list the hyperparameters of DAgger training in Table~\ref{table:dagger_hype}. Moreover, we randomize the camera viewpoint to account for discrepancies between the simulated and real-world camera poses. The action head is a three-layer MLPs with two hidden layers using ReLU activations; its outputs are subsequently squashed with a tangent activate function to enforce bounded actions in the target action space. 
\begin{table}[h]
    \centering
    \caption{DAgger Training Hyperparameters}
    \resizebox{0.4\textwidth}{!}{\begin{tabular}{c|c}
    \hline
    Description & Value \\
    \hline
        Buffer size  & $1e^{5}$ \\
        Learning rate  & $1e^{-4}$ \\
        Optimizer  & Adam \\
        Scheduler adjusted probability $p$ & $0.93$ \\
        Optimizer eps & $1e^{-8}$ \\
        Batch size & $256$ \\
        Action repeat & $1$ \\
        Frame stack & $3$ \\
        Depth threshold distance $\mathbf{d}$ & $0.9$ to $1.1$ \\

    \end{tabular}}
    \label{table:dagger_hype}
\end{table}

The DAgger training action objective is:
\begin{equation}
  \mathcal{L}_{\mathrm{act}}(\theta)
  = \mathbb{E}_{(\mathbf{o}_t,\mathbf{a}^{\star}_t)\sim\mathcal{D}}
    \Big[\,\|\mathbf{a}_t-\mathbf{a}^{\star}_t\|_2^{2}\,\Big]
  + \mathbb{E}_{(\mathbf{o}_t,\mathbf{a}^{\star}_t)\sim\mathcal{D}}
    \Big[\,\|\mathbf{a}_t-\mathbf{a}^{\star}_t\|_1\,\Big],
  \label{eq:act_loss_expect}
\end{equation}

where $\mathbf{a}_t=\pi_{\theta}(\mathbf{o}_t)$, and $\mathbf{a}^{\star}_t$ is the expert action.  

\subsection{Navigation Details}
\label{sec:nav}
We adopt the same navigation configuration as employed in ViNT. For localization, we first capture RGB-D images of the target position. Then, during navigation, the images recorded along the reference trajectories serve as goal observations for inferring the subsequent actions. In addition, we configure the number of waypoints to $4$ and set the radius to $5$, tailoring these parameters to the operational characteristics of our mobile platform.

\subsection{Instance Generalization}
\begin{figure}[t]
  \centering
  \includegraphics[width=1.0\linewidth]{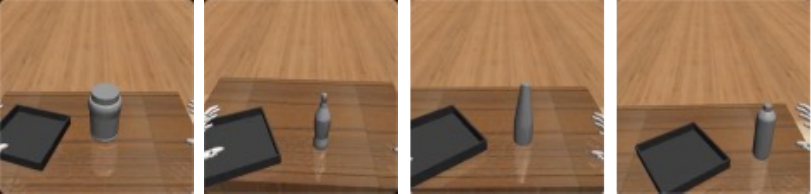}
  \caption{ \textbf{Object generalization visualization.} To help the policy adapt to different object shapes, we randomize each object’s geometry during training to prompt the policy to adjust its actions accordingly.}
  \label{fig:obj_generalization}
\end{figure}
We also conduct experiments to assess the object-level generalization capability of our approach. For the handover task, we select several bottles from the UniDexGrasp++ dataset~\cite{wan2023unidexgrasp++} and introduce randomized variations in bottle shapes during each environment reset. The corresponding visualizations are presented in Figure~\ref{fig:obj_generalization}. The trained policy, distilled via the DAgger framework, demonstrates zero-shot generalization to bottles of varying shapes in real-world settings.

\begin{figure*}[t]
  \centering
  \includegraphics[width=1.0\linewidth]{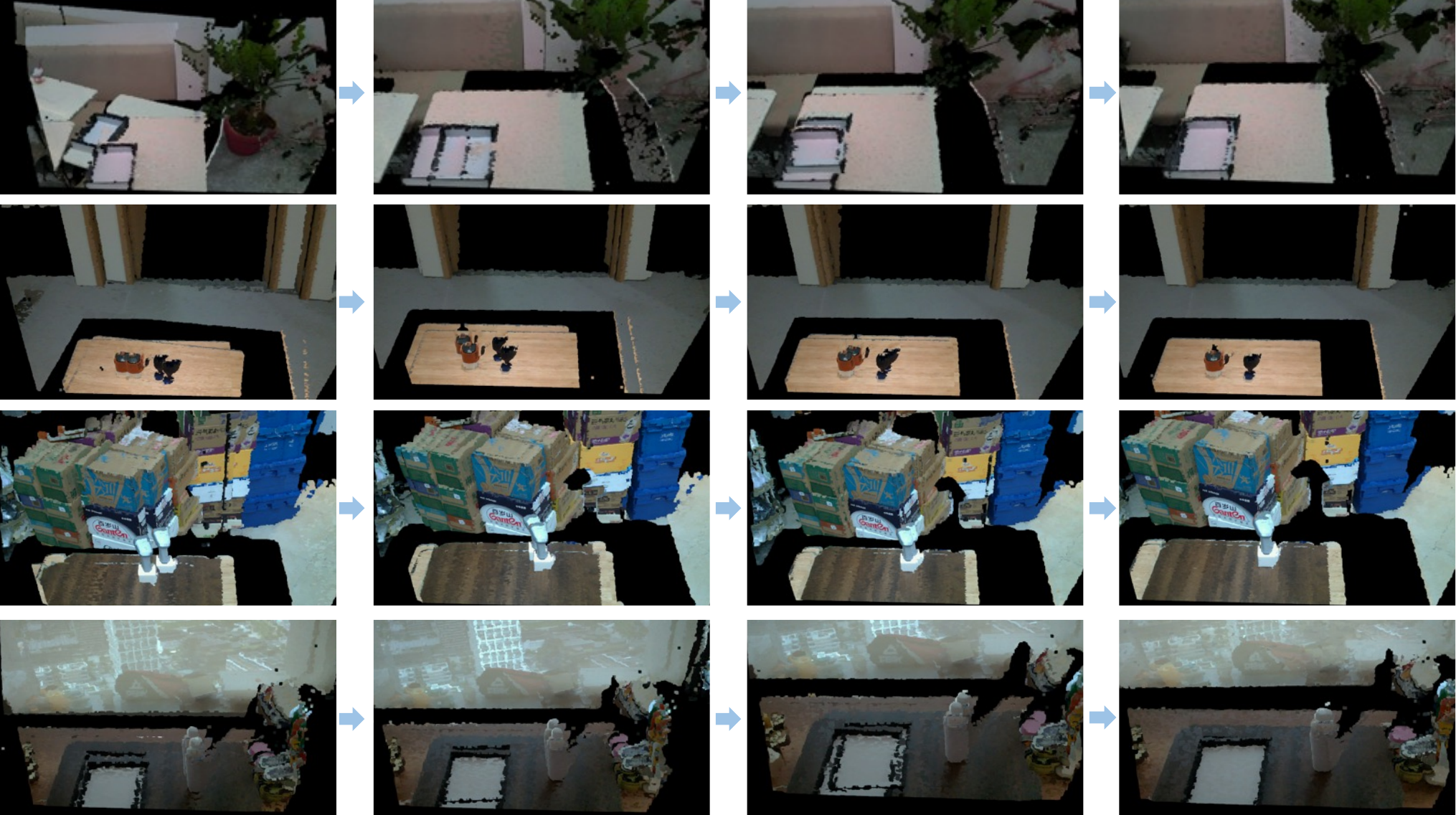}
  \caption{\textbf{Closed-loop PnP visualization.} Across diverse scenarios, the closed-loop PnP algorithm iteratively refines the robot’s pose and ultimately aligns it with the desired target pose with high precision.}
  \label{fig:closed-loop}
\end{figure*}

\subsection{Qualitative Analysis of Closed-loop PnP}

Figure~\ref{fig:closed-loop} illustrates that across various scenarios, the closed-loop PnP localization with iterative refinements progressively aligns the captured point clouds with the target positions, underscoring the localization precision achieved by \ourshort.

\subsection{DAgger Training}
\label{sec:dagger_train}
\begin{figure}[h]
  \centering
  \includegraphics[width=1.0\linewidth]{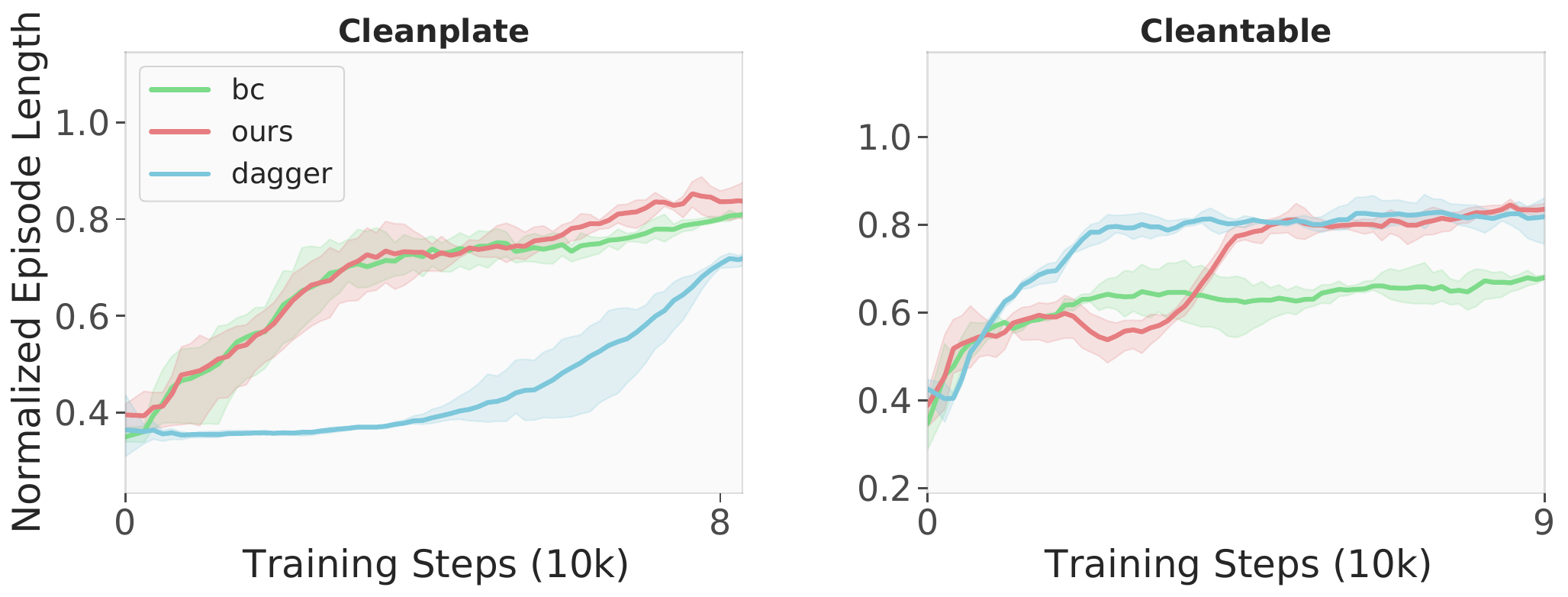}
  \caption{ \textbf{DAgger training efficiency.} \ourshort attains high sample efficiency and asymptotic performance across different types of tasks.}
  \label{fig:dagger_curves}
\end{figure}

Regarding DAgger training, we compare our implementation of DAgger with both pure expert training~(i.e., Behavior Cloning) and pure student training. Figure~\ref{fig:dagger_curves} reports each method’s performance on two representative tasks. \textit{Clean Table} is a long-horizon task in which the robot must first place two objects into a box and then move the box to the center of the table. As shown in Figure~\ref{fig:dagger_curves}, pure imitation learning struggles to cover long-horizon and out-of-distribution states. In \textit{Clean Plate}, where the human motion is extracted from raw videos, the expert policy’s actions contain non-negligible noise; consequently, during early interactive training, action errors can be amplified by querying this noisy expert, leading to poor sample efficiency. In contrast, HERMES warms up the policy with expert data and then gradually reduces reliance on the expert action distribution, achieving high sample efficiency without sacrificing asymptotic performance in both tasks.

\subsection{Hybrid Control for Real-world Evaluation}
\label{sec:hybrid}
\begin{figure}[t]
  \centering
  \includegraphics[width=0.9\linewidth]{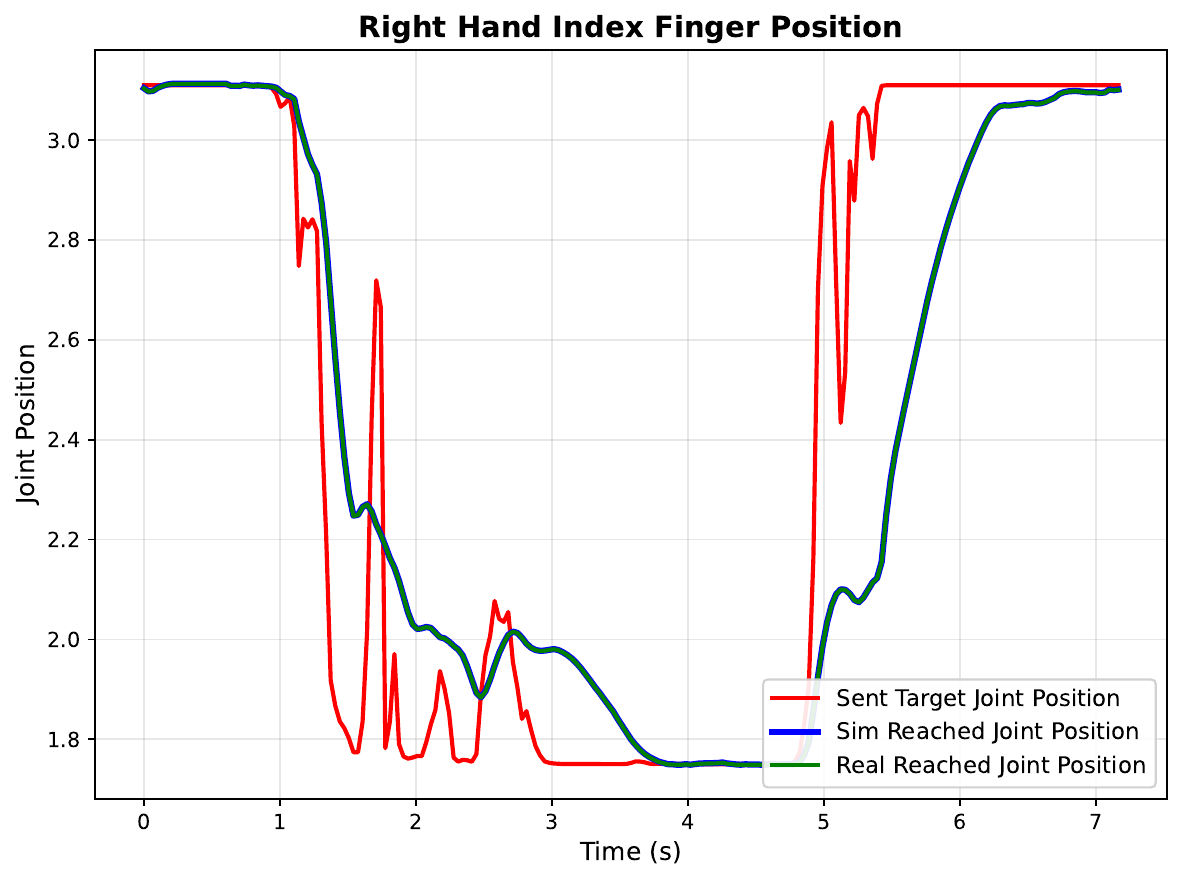}
  \caption{ \textbf{The comparison of sent target joint position and the actual reached joint position in both simulation and real-world.} The red, blue, and green curves exhibit similar overall trends. Compared with the blue and green curves, the red curve shows a deviation in joint position at the same time step, while the blue and green traces overlaps throughout. This discrepancy points to the gap between the simulated and real-world dynamics, and adopting the hybrid control keeping the dynamics consistency between the simulation and the real robot.}
  \label{fig:hybrid-control}
\end{figure}
It is worth noting that, in our hybrid sim2real control framework, the simulation environment does not contain any object models. We rely on the simulated robot to carry out all kinematic and dynamic computations and to maintain alignment with the real system. Furthermore, to showcase the benefits of our hybrid sim2real control strategy, we visualize the temporal trajectory of the right-hand index-finger joint angle during policy execution. Figure~\ref{fig:hybrid-control} shows a pronounced, nonlinear discrepancy between the joint target position inferred by the policy and the actual reached joint position. Such a sizable mismatch is difficult for the policy to compensate for. In contrast, the hybrid sim2real control strategy forces the simulated and real robots to share an identical dynamics model. As demonstrated in Figure~\ref{fig:hybrid-control}, the blue and green traces remain consistently overlapped across the entire horizon, indicating that hybrid control substantially reduces the sim2real gap.

\subsection{The Sequential Adjustment Strategy of Closed-loop PnP}

During the closed-loop PnP stage, the system takes a goal RGBD image $I_g$ and a current robot-captured RGBD image $I_c$ as inputs. At each timestep $t$, the robot updates $I_c$ with its current view. The $\text{PnP}$ algorithm which is made up of RANSAC PnP \cite{zuliani2009ransac} and refine PnP \cite{madsen2004methods,eade2013gauss,bradski2000opencv} computes the relative rotation $\mathbf{R} \in \mathbb{R}^{3\times 3}$ and translation $\mathbf{t} \in \mathbb{R}^3$ between the current and goal poses. The pose errors in x direction, y direction and yaw orientation are derived as $e_{\text{x}}$, $e_{\text{y}}$, $e_{\text{yaw}}$ from the relative rotation and translation. These errors drive the PID controller \cite{willis1999proportional} to output $v_{\text{x}}$, $v_{\text{y}}$, $v_{\text{yaw}}$, which move the robot base to reduce the errors. The adjustment loop terminates when all errors satisfy $|e_j| < \varepsilon_j$ for $j \in {\text{x, y, yaw}}$.

\begin{algorithm}
\caption{Closed-loop PnP Adjustment}\label{alg:PnP}
\begin{algorithmic}
\Require Goal RGBD image $I_g$, thresholds $\varepsilon_{\text{x}}$, $\varepsilon_{\text{y}}$, $\varepsilon_{\text{yaw}}$
\Ensure Adjusted robot pose
\State Initialize PID controllers
\While{$\max\left( \frac{|e_{\text{x}}|}{\varepsilon_{\text{x}}}, \frac{|e_{\text{y}}|}{\varepsilon_{\text{y}}}, \frac{|e_{\text{yaw}}|}{\varepsilon_{\text{yaw}}} \right) > 1$}
    \State Capture current RGBD image $I_c$
    \State $\mathbf{R}, \mathbf{t} \gets \text{PnP}(I_g, I_c)$ \Comment{Using RANSAC + Refine PnP}
    \For{$j \in \{\text{x, y, yaw}\}$}
        \State $e_j  \gets\text{ExtractPoseErrors}(\mathbf{R,T})$
        \State $v_j \gets \text{PIDController}(e_j)$
    \EndFor
    \State Send velocity command $(v_{\text{x}}, v_{\text{y}}, v_{\text{yaw}})$ to robot
\EndWhile
\end{algorithmic}
\end{algorithm}

\begin{figure*}[t]
  \centering
  \includegraphics[width=1.0\linewidth]{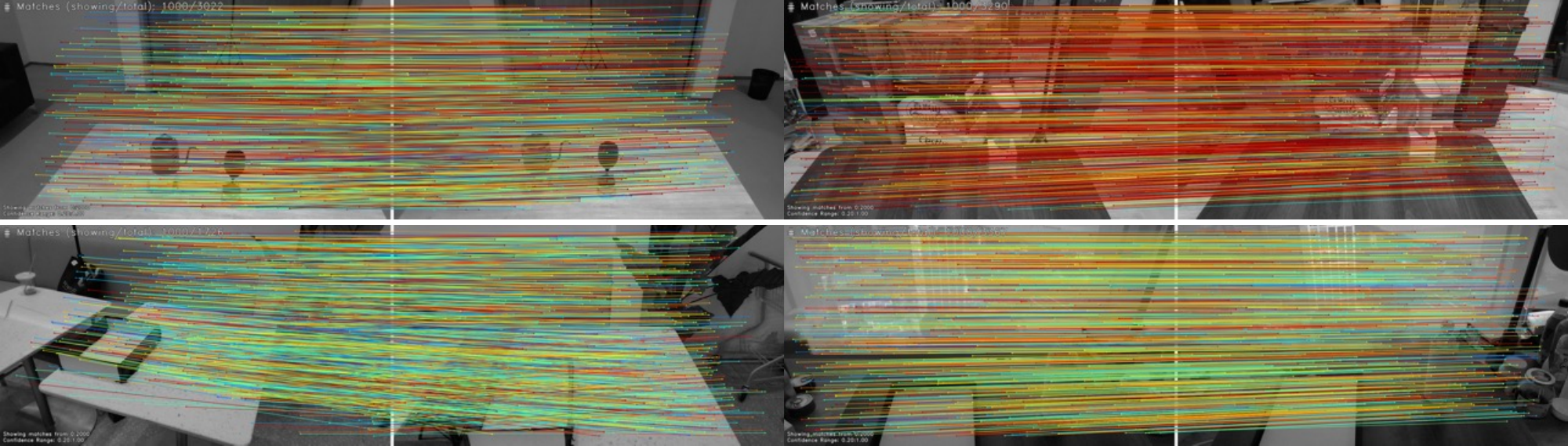}
  \caption{ \textbf{Feature matcher visualization of Efficient LoFTR.} Efficient LoFTR is capable of establishing correspondences between the target RGB image and the current RGB frame at a high frequency.}
  \label{fig:feature}
\end{figure*}
\subsection{Feature Matcher Visualization}

Efficient LoFTR~\cite{wang2024efficient} consistently extracts dense feature correspondences at high frequency across various environments. This capability provides the rich feature correspondence required for the subsequent PnP pose estimation. The visualizations are shown in Figure~\ref{fig:feature}.

\subsection{Comparison with the depth image from Depth Anything }
\begin{figure}[t]
  \centering
  \includegraphics[width=0.9\linewidth]{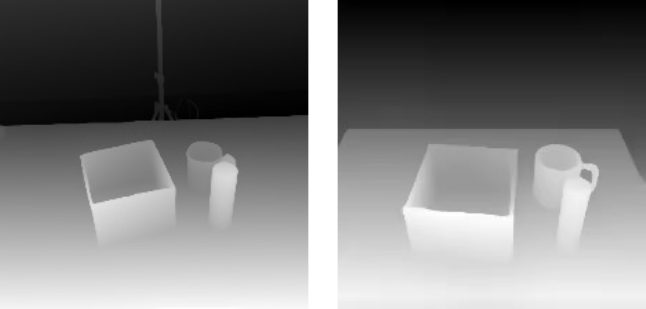}
  \caption{ \textbf{Comparison of Depth-anything generated images between simulation and real-world.} Depth-Anything can produce depth images that are semantically aligned between simulation and real-world.}
  \label{fig:dpt-compare}
\end{figure}
\begin{figure}[h]
  \centering
  \includegraphics[width=1.0\linewidth]{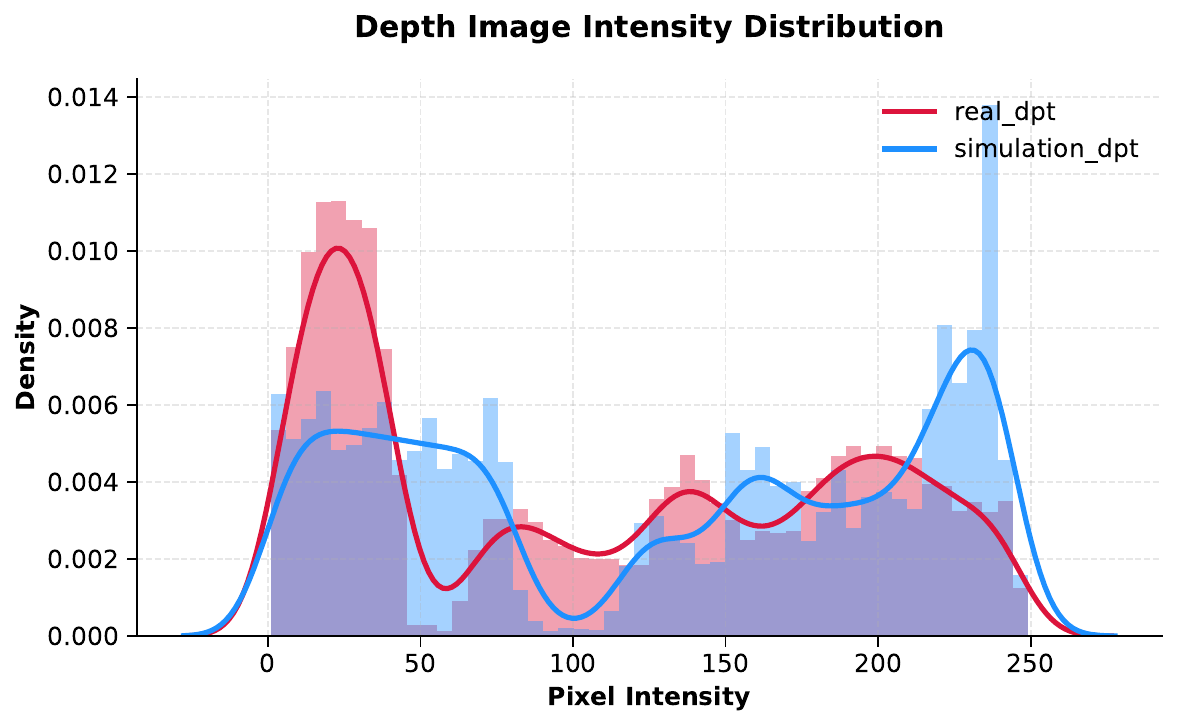}
  \caption{\textbf{Depth intensity distribution of Depth-anything generated images. }The red curve shows the depth-value distribution extracted from real-world RGB images, whereas the blue curve corresponds to that obtained from simulated images. Despite their semantic alignment, the depth maps estimated by Depth-Anything reveal a pronounced quantitative gap between the simulated and real-world domains.}
  \label{fig:dpt_dist}
\end{figure}
Depth-Anything~\cite{yang2024depth,yang2024depth2}, a widely adopted foundation model for depth estimation, demonstrates remarkable robustness and fine-grained detail perception, even under diverse and unstructured real-world conditions. Building on these capabilities, we also leverage Depth-Anything to bridge the perceptual gap between simulation and the real world. Consistent with \ourshort, we also apply the \textit{mixup} augmentation strategy with the disparity map to enhance the robustness. We first utilize the NYU Depth Dataset~\cite{silberman2012indoor} and transform the depth maps $D$ into disparity maps $\hat{I}$ to better align with the  distribution of DA generated images:
\begin{equation}
    \hat{I}(x, y) = \frac{1}{D(x, y)},
\end{equation}
where $(x, y)$ is the pixel position.

We then blend the DA input $\boldsymbol{o}_{DA}$ with a certain disparity map $\boldsymbol{o}_{disparity}$: ${\boldsymbol{\hat{o}}} = \alpha*\boldsymbol{o}_{DA} + (1 - \alpha)*\boldsymbol{o}_{disparity}$, where $\alpha$ is the coefficient. ${\boldsymbol{\hat{o}}}$ is then used as the final visual input to the network.

We compare our  depth-augmentation scheme with the depth maps synthesized by Depth-anything-v2. As reported in Table~\ref{table: compare_da}, \ourshort achieves better performance and success rates. Although Figure~\ref{fig:dpt-compare} shows that the synthetic depth maps can achieve semantic coherence between simulation and real-world depth estimation, Figure~\ref{fig:dpt_dist} reveals a distributional discrepancy in the underlying pixel-wise depth estimates across the two domains. This numerical gap ultimately suppresses overall success rates, making the method less effective than our direct preprocessing of raw depth images.

\begin{table}[h]
\centering
\caption{Comparison of \ourshort and Depth-Anything}
\label{table: compare_da}
\resizebox{0.4\textwidth}{!}{
\begin{tabular}{c|cc}
\toprule

\begin{tabular}[c]{@{}c@{}} \multicolumn{1}{c}{\multirow{1}{*}{Tasks $\backslash$ Method}} \end{tabular}   & \multicolumn{1}{c}{\ourshort} & \multicolumn{1}{c}{Depth-Anything} \\

\midrule

Bottle Handover & \best{66.7} & $40.0$  \\

Scan Bottle &	\best{73.3} & $33.3$  \\

\bottomrule
\end{tabular}
}
\end{table}

\end{document}